\pdfoutput=1
\RequirePackage{fix-cm}
\PassOptionsToPackage{table,xcdraw}{xcolor}
\PassOptionsToPackage{capitalise}{cleveref}
\documentclass[10pt,letterpaper]{template}
\usepackage[utf8]{inputenc}
\usepackage{siunitx,tikz,pgfplots,enumitem,makecell,pifont}
\usepackage{tabularx,longtable}
\usepackage[ruled,vlined]{algorithm2e}
\usepackage{appendix}
\pgfplotsset{compat=1.17}

\usepackage{amsmath,amsfonts,bm}

\def\eqref#1{equation~\ref{#1}}

\def\1{\bm{1}}

\newcommand{\valid}{\mathcal{D_{\mathrm{valid}}}}

\DeclareMathAlphabet{\mathsfit}{\encodingdefault}{\sfdefault}{m}{sl}
\SetMathAlphabet{\mathsfit}{bold}{\encodingdefault}{\sfdefault}{bx}{n}

\crefname{algorithm}{Algorithm}{Algorithms}
\Crefname{algorithm}{Algorithm}{Algorithms}
\AddToHook{cmd/appendix/before}{%
  \crefalias{section}{appendix}%
  \crefalias{subsection}{appendix}%
  \crefalias{subsubsection}{appendix}%
}

\newcommand{\algo}{RoboAware}
\newcommand{\pschema}{\ensuremath{\mathrm{P}^{5}}}
\newcommand{\scb}{\textsc{SCB}}
\newcommand{\eal}{\textsc{EAL}}
\newcommand{\familyset}{\ensuremath{\mathcal{C}}}
\newcommand{\modfamily}{\ensuremath{\mathrm{Mod}}}
\newcommand{\eetfamily}{\ensuremath{\mathrm{E2E}}}
\newcommand{\codepolicy}{\ensuremath{\pi_{\mathrm{code}}}}
\newcommand{\performendtoend}{\texttt{Perform\_E2E}}

\newcommand{\imageOn}{1}
\newcommand{\figimgplaceholder}[3]{%
  \fbox{\parbox[c][#2][c]{#1}{\centering #3}}%
}
\newcommand{\figimg}[4]{%
  \ifnum\imageOn=1
    \IfFileExists{#2}{\includegraphics[width=#1]{#2}}{\figimgplaceholder{#1}{#3}{#4}}%
  \else
    \figimgplaceholder{#1}{#3}{#4}%
  \fi
}

\newcommand{\NA}{\multicolumn{1}{c}{\textemdash}}
\newcommand{\NR}{\multicolumn{1}{c}{\textit{n.r.}}}

\crefname{algocf}{Algorithm}{Algorithms}
\Crefname{algocf}{Algorithm}{Algorithms}

\renewcommand{\surveymaintitle}{RoboAware: Learning to Coordinate Embodied Skills\\from Counterfactual Outcomes}
\renewcommand{\surveysubtitle}{}
\renewcommand{\surveyrunningtitle}{RoboAware: Learning to Coordinate Embodied Skills from Counterfactual Outcomes}
\renewcommand{\surveyauthors}{%
  {\normalfont\rmfamily\bfseries\fontsize{9.6}{12}\selectfont\color{ink}%
  \resizebox{\linewidth}{!}{%
  \begin{tabular}{@{}c@{}}
  Bohan Zhou\textsuperscript{1*\dag},
  Xingbei Chen\textsuperscript{2,*},
  Emily Huang\textsuperscript{3},
  Weilin Ruan\textsuperscript{1},
  Haojian Huang\textsuperscript{4},
  Yehang Zhang\textsuperscript{2},
  Zexi Li\textsuperscript{1}\\
  Wenqian Li\textsuperscript{3},
  Qize Yu\textsuperscript{5},
  Zetian Song\textsuperscript{5},
  Leyi Wu\textsuperscript{2},
  Jinghao Li\textsuperscript{1},
  Mingxuan Song\textsuperscript{5},
  Xinrun Xu\textsuperscript{6}\\
  Zongyang Qiu\textsuperscript{2},
  Yangkai Wei\textsuperscript{3},
  Tianyi Zhang\textsuperscript{3},
  Kaiwen Zhou\textsuperscript{3},
  Yinchuan Li\textsuperscript{3},
  James Cheng\textsuperscript{1}
  \end{tabular}}\par
  \vspace{2pt}%
  {\normalfont\rmfamily\mdseries\fontsize{8}{9.6}\selectfont
  \textsuperscript{*}Equal Contribution.\quad
  \textsuperscript{\dag}Project Leader.\par}}}
\renewcommand{\surveyaffiliations}{%
  {\normalfont\rmfamily\fontsize{8}{9.6}\selectfont\color{ink}%
  \textsuperscript{1}The Chinese University of Hong Kong\quad
  \textsuperscript{2}The Hong Kong University of Science and Technology\quad
  \textsuperscript{3}Knowin AI\\
  \textsuperscript{4}The University of Hong Kong\quad
  \textsuperscript{5}Peking University\quad
  \textsuperscript{6}University of the Chinese Academy of Sciences\par}}
\renewcommand{\surveycorrespondence}{%
  {\normalfont\sffamily\fontsize{9.2}{12}\selectfont\color{ink}%
  \raisebox{-.17em}{\includegraphics[height=1.05em]{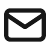}}\hspace{.5em}%
  {\fontsize{8}{10}\selectfont\bfseries CONTACT}\hspace{.9em}%
  {\normalfont\ttfamily\fontsize{9.2}{12}\selectfont
    \{\href{mailto:zhoubh@link.cuhk.edu.hk}{zhoubh}\}}\hspace{.15em}%
  {\color{muted}@}\hspace{.3em}%
  \tcbox[on line,colback=ink,colframe=ink,boxrule=0pt,arc=2pt,
    boxsep=0pt,left=5pt,right=5pt,top=2pt,bottom=2pt]{%
    \normalfont\sffamily\bfseries\fontsize{9}{10}\selectfont\color{white}link.cuhk.edu.hk}%
  \par}}
\renewcommand{\surveyinstitutionlogos}{}
\renewcommand{\surveyresourcelinks}{}
\renewcommand{\surveycontactdetails}{}
\renewcommand{\surveyfooternote}{}
\fancypagestyle{firststyle}{%
  \fancyhf{}\fancyfoot[C]{\normalfont\small\color{muted}\thepage}%
  \renewcommand{\headrulewidth}{0pt}}

\title{RoboAware: Learning to Coordinate Embodied Skills from Counterfactual Outcomes}
\author{Bohan Zhou, Xingbei Chen, Emily Huang, Weilin Ruan, Haojian Huang, Yehang Zhang, Zexi Li, Wenqian Li, Qize Yu, Zetian Song, Leyi Wu, Jinghao Li, Mingxuan Song, Xinrun Xu, Zongyang Qiu, Yangkai Wei, Tianyi Zhang, Kaiwen Zhou, Yinchuan Li, James Cheng}
\hypersetup{
  pdftitle={RoboAware: Learning to Coordinate Embodied Skills from Counterfactual Outcomes},
  pdfauthor={Bohan Zhou, Xingbei Chen, Emily Huang, Weilin Ruan, Haojian Huang, Yehang Zhang, Zexi Li, Wenqian Li, Qize Yu, Zetian Song, Leyi Wu, Jinghao Li, Mingxuan Song, Xinrun Xu, Zongyang Qiu, Yangkai Wei, Tianyi Zhang, Kaiwen Zhou, Yinchuan Li, James Cheng},
  breaklinks=true,
  bookmarksdepth=3
}
\abstract{
Embodied coding agents can combine modular robot skills with frozen end-to-end policies, yet effective composition requires anticipating which policy family will succeed in the current physical state. We present \textbf{\algo{}}, which builds on coding agents' skill orchestration by learning only a state-conditioned responsibility coordinator from counterfactual outcomes. Inspired by the success of REPL, we propose the \textbf{\pschema{}} schema and formulate a hierarchical MDP based on it. \pschema{} organizes skills uniformly into five semantic stages, defining where responsibility can be compared. To address the lack of counterfactual branch outcomes in existing work, we introduce \textbf{State-Locked Counterfactual Branching (SCB)}, which restores the same training state to generate and execute a code block from each admissible family, exposing outcomes that selected-branch experience leaves unobserved. Building on this, we propose \textbf{Execution-Aware Learning (EAL)}, which combines Monte Carlo tree search with Q-learning to distill these outcomes into family-conditioned values. At deployment, the coordinator selects the policy family according to observable context, and the frozen coding agent generates the next local code block. Comprehensive single-episode evaluations on 100 tasks show that \algo{} reaches a \textbf{77.0\%} overall success rate, with SOTA averages of 90.0\% on RoboSuite, 73.8\% on diverse LIBERO-Pro task clusters, and 90.0\% on challenging RoboTwin bimanual tasks, outperforming existing code-as-policy and VLA-harness baselines.

}

\begin{document}
\maketitle
\pagestyle{surveybody}
\begin{figure}[hb]
	\centering
		\figimg{\linewidth}{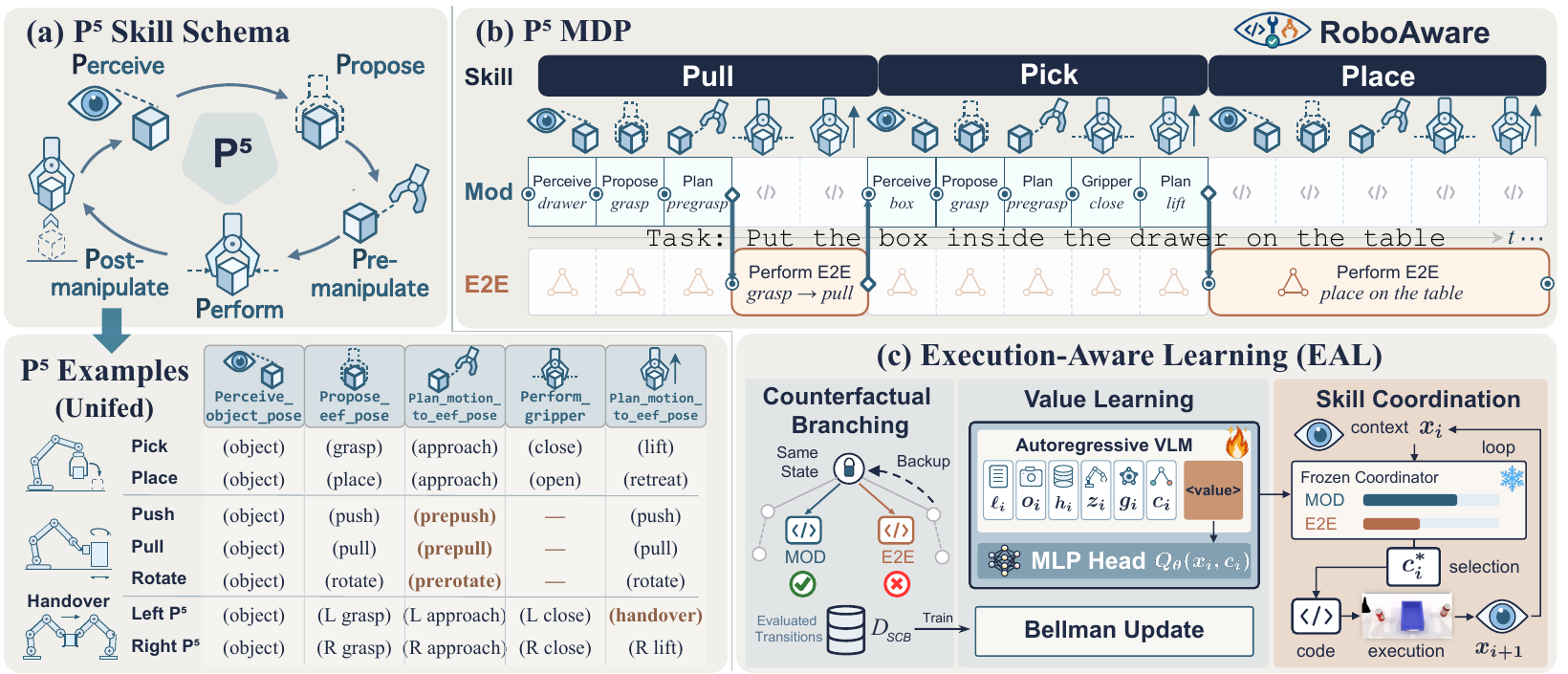}{6.5cm}{%
			[(a) \pschema{} skill schema]\\[0.6em]
			[(b) \pschema{} MDP: $c_i\!\rightarrow\!a_i$]\\[0.6em]
			[(c) \eal{}: \scb{} $\rightarrow Q_\theta$ $\rightarrow$ coordination]}
		\caption{Overview of \algo{}.
		(a)~The \pschema{} skill schema decomposes each skill into five stages---perceive, propose, pre-manipulate, perform,
    and post-manipulate---and unifies primitive skills (\textit{e.g.}, pick, place, push, pull) and compositional skills (\textit{e.g.}, handover) under one representation.
		(b)~A hierarchical MDP selects a policy family $c_i\in\{\modfamily,\eetfamily\}$ at each \pschema{} stage decision boundary; a coding agent then generates each corresponding code block $a_i$.
		(c)~\eal{} learns family-conditioned coordination: \scb{} evaluates modular and end-to-end branches from the same restored state into $\mathcal{D}_{\mathrm{SCB}}$, Q-learning yields $Q_\theta(x,c)$, and at test time
    a frozen coordinator selects best $c_i$ to condition code generation and execution.
  }
	\label{fig:overview}
\end{figure}

\section{Introduction}

Performing general-purpose embodied manipulation in diverse and unstructured environments remains challenging in the absence of low-level executor adaptation.
Recent progress toward this goal has followed two major paradigms, namely end-to-end policies and coding agents, each with distinct strengths and limitations.
Within the end-to-end paradigm, task-specific visuomotor controllers excel at contact-rich control but typically lack native support for open-vocabulary rebinding~\citep{ACT,DiffusionPolicy,DP3}. Vision-language-action (VLA) and world-action models (WAMs) support language-conditioned execution~\citep{RT-1,RT-2,OpenVLA,pi0}, yet become unreliable under compositional and state-distribution shifts.
Coding agents, by contrast, enable Code-as-Policy~\citep{CaP,Cap-X,VoxPoser} by flexibly composing program modules for long-horizon tasks. Their physical competence, however, is bounded by the exposed primitives, limiting fine-grained, closed-loop correction during contact-rich interactions.
These complementary advantages inspire hybrid frameworks in which coding agents generate modular API compositions or invoke frozen end-to-end policies, most commonly VLAs, as contact-rich primitives~\citep{HarnessVLA,HVLA,RoboHarness}.
Nevertheless, high-level coding agents lack inherent awareness of the divergent consequences of invoking different skill executors.
They struggle to choose among an explicit modular composition, any available self-contained policy, and their possible switching patterns.
For instance, modular execution can remain stable under task-distribution shift, whereas end-to-end execution excels in case of unexpected interruption.
This raises a critical research question: \textit{How can we endow coding agents with outcome-aware execution responsibility over embodied skills?}

We address this question with \textbf{\algo}, a framework that learns a responsibility boundary between modular composition and end-to-end execution from counterfactual outcomes. 
Here, a \emph{policy family} $c\in\{\modfamily,\eetfamily\}$ is the sole high-level coordination action; the \emph{API library} comprises typed APIs together with their bound low-level backends; \emph{responsibility} denotes which family should control the next semantically meaningful skill transition. 
The first challenge is to define where responsibility can be compared.
The space of low-level code generation is too expansive and unstructured to provide a tractable basis for responsibility assignment.
Inspired by the success of REPL-style programming~\citep{REPL-Plan}, we propose the \textbf{$\mathbf{P}^5$ Skill Schema} to address this limitation.
The \pschema{} schema organizes embodied skills into semantically meaningful stages from perception to execution in 3D space.
We formulate a finite-horizon hierarchical MDP that selects a policy family at each \pschema{} stage.
At the lower level, we curate an API library combining typed modular interfaces for composability and controllability with end-to-end policies for contact-rich local execution.
Conditioned on the selected family and current context, the frozen coding agent $\codepolicy$ generates a complete local code block through the fixed API library.
Learning is confined to the responsibility coordinator, while skill implementation leverages the coding agent's existing code-generation capability.

The second challenge is to expose the missing outcome of the skills that were not selected.
Many approaches execute open-loop skill sequences or code programs without attending to outcomes~\citep{CaP,ProgPrompt,VoxPoser}.
Methods that obtain outcomes typically rely on VLM-based evaluations limited to the predicted outcome of the executed trajectory, and therefore inherit selection bias~\citep{RoboHarness,VLAProbe}.
We introduce \textbf{State-Locked Counterfactual Branching (\scb{})}, which leverages simulator (1) save-restore mechanisms and (2) oracle outcome judgments to execute every admissible family from an identical physical state, producing a same-state set of potential outcomes at each \pschema{} decision point.
To convert these set-wise outcomes into deployable family selection, we formulate \textbf{Execution-Aware Learning (\eal{})}, in which Monte Carlo tree search (MCTS) treats restorable \pschema{} decision states as nodes and admissible policy families as edges, while Q-learning distills the explored outcomes into a family-conditioned value function $Q_\theta$.
The family-wise values encode a relative responsibility preference over modular composition and the end-to-end family, reflecting expected remaining skill- and task-level return under an observation.
During deployment, the \algo{} coordinator assigns responsibility by maximizing $\rho_\theta$ induced by $Q_\theta$, and a coding agent generates executable code blocks accordingly to transition from one \pschema{} state to the next, without state restoration or privileged simulator state.
See \cref{fig:overview} for an overview of the proposed framework. Our contributions are as follows:

\begin{itemize}[leftmargin=*]
		\item We formulate \algo{} as a hierarchical MDP that selects a policy family $c_i$ at each \pschema{} stage to produce code block $a_i$. We curate its lower-level API library around two complementary families, namely composable modular components and end-to-end policies for contact-rich local skills.
		\item We introduce \scb{} to acquire paired locked-state outcomes, and \eal{} to learn family-conditioned $Q_\theta$ through MCTS-guided exploration and Q-learning. 
    \item We recover the coordinator from $Q_\theta$ and evaluate it on 100 tasks across three benchmarks, obtaining a SOTA 77.0\% overall success rate. Ablations confirm the effectiveness of our designs.
\end{itemize}

\section{Methodology}
\subsection{Problem Formulation}
We consider language-conditioned manipulation in a physics environment $\mathcal{E}$, where the observation at time $t$ is $o_t=(I_t^{\mathrm{rgb}}, I_t^{\mathrm{d}}, q_t)$ from an agent-view RGB image, a metric depth map, and proprioception.
We formulate the task as a finite-horizon MDP $\mathcal{M}=(\mathcal{X},\familyset,\mathcal{P},r,\gamma)$.
Decision point $i$ occurs at the start of a \pschema{} stage, where the observation $x_i=(\ell,o_i,h_i,z_i,g_i)\in\mathcal{X}$ is received. Here $\ell$ denotes the language instruction, $h_i$ the observable execution history, $z_i$ the current \pschema{} stage, and $g_i$ the active skill schema.
The high-level action is a policy-family choice $c_i\in\mathcal{C}_i(z_i,g_i)\subseteq\familyset=\{\modfamily,\eetfamily\}$.
A coding agent generates a complete executable code block $a_i\sim\codepolicy(\cdot\mid x_i,c_i)$ with $a_i\in\mathcal{A}_{c_i}(x_i,z_i,g_i)$, where $\mathcal{A}_{\mathrm{Mod}}$ and $\mathcal{A}_{\mathrm{E2E}}$ are the modular- and end-to-end-family code-block spaces.
Executing $a_i$ in $\mathcal{E}$ until the next decision point induces the transition $\mathcal{P}(x_{i+1}\mid x_i,c_i)=\mathbb{E}_{a_i\sim\codepolicy(\cdot\mid x_i,c_i)}[\mathcal{P}_{\mathcal{E}}(x_{i+1}\mid x_i,a_i)]$.
A binary reward $r$ is received at task termination. $\gamma=1$.
Only the coordinator's value model $Q_\theta$ is trained; the coding agent, API interfaces, backend bindings, and low-level backend parameters remain fixed.
\subsection{\algo{} Framework}
\begin{figure}[htbp]
  \centering
  \figimg{\linewidth}{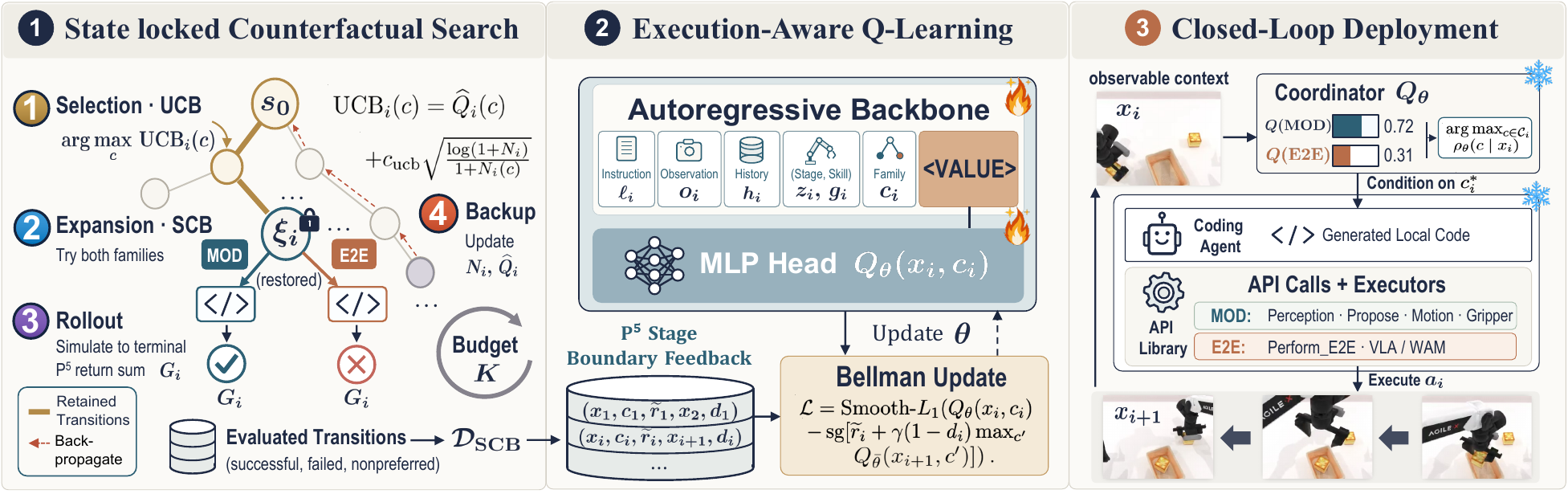}{4.8cm}{%
      [(1) State-locked counterfactual search]\\[0.6em]
      [(2) Execution-aware Q-learning]\\[0.6em]
      [(3) Closed-loop deployment]}
  \caption{Method overview of \eal{} in \algo{}.
  (1)~State-locked counterfactual search repeats Selection (UCB), Expansion (\scb{} paired \modfamily{}/\eetfamily{} branches from a restored $\xi_i$), Rollout (suffix return $G_i$), and Backup within budget $K$, retaining successful, failed, and nonpreferred transitions in $\mathcal{D}_{\mathrm{SCB}}$.
  (2)~A VLM backbone with an MLP head maps context--family tokens ending in \texttt{<VALUE>} to $Q_\theta(x_i,c_i)$, trained by Bellman updates on skill-completion and terminal-task feedback.
  (3)~At deployment, frozen $Q_\theta$ induces $\rho_\theta$ and selects $c_i^*=\arg\max_c\rho_\theta(c\mid x_i)$; a frozen coding agent generates a local code block against the fixed API library and executes it, looping on observable feedback at each \pschema{} boundary.}
  \label{fig:method}
\end{figure}
\algo{} formulates heterogeneous policy-family orchestration as stage-conditioned responsibility learning.
The
pipeline is visualized in \Cref{fig:method} and elaborated in \Cref{alg:roboaware-deploy}.
\subsubsection{Skill Preparation}
\textbf{$\mathbf{P}^{5}$ Skill Schema}.
We instantiate local embodied skills through either interface-based modular composition or a self-contained end-to-end controller.
The \pschema{} skill schema decomposes each skill into five stages: \textit{perceive}, \textit{propose}, \textit{pre-manipulate}, \textit{perform}, and \textit{post-manipulate}.
As shown in \Cref{fig:overview}, this schema provides a unified representation for arbitrary skills, including primitive skills (\textit{e.g.} pick, place, push, pull) and compositional skills (\textit{e.g.} handover).
Given the task instruction $\ell$ and initial observation $o_0$, the coding agent constructs a task-level plan $\mathcal G$ comprising \pschema{} skill schemas.
Responsibility is reconsidered at each \pschema{} stage. 

\textbf{API Library Construction}. Building on the \pschema{} stages, we expose an API library that supports the modular and end-to-end policy families.
The \modfamily{} family uses an explicit intermediate interface for composability and controllability, whereas the \eetfamily{} family relies on an implicit internal loop for local contact competence.
We assemble the library by offline screening of open-source modular perception, grasping, and motion components alongside released VLA and world-action-model (WAM) policies.
For the \modfamily{} family, screening retains \textbf{InstructSAM}~\citep{InstructSAM}, our augmented \textbf{NeuGraspNet+} from NeuGraspNet~\citep{NeuGraspNet}---which expands sparse geometric candidates and selects Top-1 only among planner-feasible grasps---and \textbf{cuRoboV2}~\citep{cuRoboV2} as the core backends for perception, pose proposal, and motion planning, respectively. 
Together with the deterministic gripper API, these components define the modular calls available to \codepolicy{}.
For the \eetfamily{} family, offline evaluation reveals limited zero-shot performance on complete tasks, yet these policies perform well on local skills such as grasping and lifting (see \Cref{sec:benchmark-performance}).
We therefore position VLA and WAM policies as local executors at grasp--lift granularity rather than as end-to-end policies for complete tasks.
\begin{table}[t]
  \begin{minipage}[t]{0.533\linewidth}
    \centering
    \captionsetup{justification=raggedright,singlelinecheck=false}
    \caption{Offline screening of the selected VLA and WAM backends (successful / total). 
    \textbf{Full}: complete task; \textbf{G--L}: grasp and lift.}
    \label{tab:vla_wam_selected_results}
    \footnotesize
    \setlength{\tabcolsep}{3.2pt}
    \resizebox{\linewidth}{!}{%
    \begin{tabular}{ccccccc}
      \toprule
      \multirow{2}{*}[-0.9ex]{Method}
        & \multicolumn{2}{c}{LIBERO-Pro}
        & \multicolumn{2}{c}{RoboSuite}
        & \multicolumn{2}{c}{RoboTwin 2.0} \\
      \cmidrule(lr){2-3}
      \cmidrule(lr){4-5}
      \cmidrule(lr){6-7}
      & Full & G--L & Full & G--L & Full & G--L \\
      \midrule
      $\pi_{0.5}$
        & \textbf{10/14} & \textbf{13/14}
        & 0/4 & \textbf{2/4}
        & 0/10 & 2/10 \\
      FastWAM
        & 4/14 & 11/14
        & 0/4 & 0/4
        & \textbf{10/10} & \textbf{10/10} \\
      \bottomrule
    \end{tabular}%
    }
  \end{minipage}%
  \hfill
  \begin{minipage}[t]{0.437\linewidth}
    \centering
    \captionsetup{justification=raggedright,singlelinecheck=false}
    \caption{High-level embodied APIs under the \pschema{} skill schema; \ding{51} marks stages covered by each API.}
    \label{tab:embodied-apis}
    \footnotesize
    \setlength{\tabcolsep}{2.2pt}
    \resizebox{\linewidth}{!}{%
    \begin{tabular}{@{}lccccc@{}}
      \toprule
      API
        & \textit{perc.}
        & \textit{prop.}
        & \textit{pre}
        & \textit{perf.}
        & \textit{post} \\
      \midrule
      \texttt{Perceive\_object\_pose}
        & \ding{51} & & & & \\
      \texttt{Propose\_eef\_pose}
        & & \ding{51} & & & \\
      \texttt{Plan\_to\_eef\_pose}
        & & & \ding{51} & & \ding{51} \\
      \texttt{Perform\_gripper}
        & & & & \ding{51} & \\
      \performendtoend
        & \ding{51} & & \ding{51} & \ding{51} & \ding{51} \\
      \bottomrule
    \end{tabular}%
    }
  \end{minipage}
\end{table}
Effectiveness and efficiency screening selects $\pi_{0.5}$~\citep{pi05} and FastWAM~\citep{FastWAM} for integration into \performendtoend{} (see \Cref{tab:vla_wam_selected_results}).
Screening details appear in \Cref{app:embodied-apis}.
Each stage admits both families except the \textit{propose} stage (see \Cref{tab:embodied-apis}).
\performendtoend{} spans the \textit{perceive}, \textit{pre-manipulate}, \textit{perform}, and \textit{post-manipulate} stages.
A call remains atomic with respect to family selection until it meets its registered postcondition, reports failure, or exhausts its step budget.
The API library, including its typed interfaces, backend bindings, and checkpoints, remains frozen during training and evaluation.

\subsubsection{Execution-Aware Learning}
Existing coding agents cannot reliably anticipate the physical consequences of executing a generated skill.
To address this limitation, we propose Execution-Aware Learning (\eal{}) as illustrated in \Cref{fig:method}, which learns family-conditioned outcome values from controlled simulator experience.

\textbf{State-Locked Counterfactual Branching} (SCB) is the training-time operator that makes the otherwise missing policy-family outcomes observable.
On a fixed set of simulator training tasks, \eal{} builds a Monte Carlo search tree from each sampled \pschema{} stage-entry context.
Nodes are restorable decision states, and outgoing edges correspond to admissible policy families.
Each node stores a snapshot $\xi_i=(x_i,m_i,p_i,r)$ containing the simulator and execution state required for exact restoration.
Beyond the deployment-visible context $x_i$, it provides oracle masks $m_i$, ground-truth object poses $p_i$, and access to the terminal reward evaluator $r$.
These privileged signals support the training rewards including pose-based assessment under threshold $\epsilon_p$.
The search repeats four search phases within a budget of $K$ completed suffix simulations per root. See \Cref{app:deployment-algo} for details.

\emph{Selection.}
Starting from the root, the search follows stored tree paths by selecting family edges with UCB at already expanded nodes, where $N_i$ and $N_i(c)$ are node and family-edge visit counts, and $\widehat{Q}_i(c)$ is the empirical return estimate:
\begin{equation}
\label{eq:ucb}
c_i^{\mathrm{tree}}=\arg\max_{c\in\mathcal{C}_i}\left[\widehat{Q}_i(c)+c_{\mathrm{ucb}}\sqrt{\frac{\log(1+N_i)}{1+N_i(c)}}\right].
\end{equation}

\emph{Expansion.}
At a new node, \scb{} evaluates every admissible family at least once from the same physical state and observable-history prefix.
For each $c\in\mathcal{C}_i$, it restores $\xi_i$, samples $a_i^c\sim\codepolicy(\cdot\mid x_i,c)$, and executes the block to the next decision point or termination.
The resulting transition contains a successor snapshot $\xi_{i+1}^c$, terminal indicator $d_i^c$, and training reward $\tilde r_i^c$.
Reevaluating an existing family edge repeats restoration, code generation, and execution with a fresh code sample.

\emph{Rollout.}
From each nonterminal successor, the codes are generated using API library to complete the suffix.
The suffix ends at task termination, unrecoverable failure, or the episode horizon $T$.
Rollout-only states are not added as expanded tree nodes during this simulation.
Each completed suffix yields $G_i=\sum_{j=i}^{T}\gamma^{j-i}\tilde r_j$.
$\tilde r_i^c$ is defined in \cref{eq:transition-reward}.

\emph{Backpropagation.}
This tree-statistics backup propagates the suffix return $G_i$ along the selected tree path, including each ancestor's recorded prefix rewards.
Each visited node and family-edge count is incremented, and $\widehat{Q}$ is updated by empirical averaging.
These statistics guide subsequent selection; the parameters of $Q_\theta$ are optimized only after data collection.
Within budget $K=20$, all evaluated tree edges, including failed and nonpreferred branches, contribute transitions to the training dataset:
\begin{equation}
\label{eq:scb-data}
\mathcal{D}_{\mathrm{SCB}}=\bigcup_i\left\{(x_i,c,\tilde r_i^c,x_{i+1}^c,d_i^c,\xi_{i+1}^c):c\in\mathcal{C}_i\right\}.
\end{equation}

\textbf{Execution-Aware Q-learning.}
A task comprises several \pschema{} skill blocks.
At the end of each \pschema{} stage, oracle signals in $\xi_{i+1}^c$ yield a perception score $r_{\mathrm{p},i}^c\in[0,1]$ from agreement with mask $m_i$, and a manipulation score $r_{\mathrm{m},i}^c\in[0,1]$ from object-pose error against $p_i$ under threshold $\epsilon_p$. The binary task-success reward $r\in\{0,1\}$ is received only when $d_i^c=1$.
The transition reward used in both tree search and Q-learning is the linear combination with non-negative weights:
\begin{equation}
\label{eq:transition-reward}
\tilde r_i^c=w_{\mathrm p}r_{\mathrm{p},i}^c+w_{\mathrm m}r_{\mathrm{m},i}^c+w_{\mathrm t}d_i^cr, \quad \text{where } w_{\mathrm p}+w_{\mathrm m}+w_{\mathrm t}=1.
\end{equation}

The learning target is
$Q_\theta(x_i,c)\approx\max_\pi\mathbb{E}_\pi\bigl[\sum_{j=i}^{T}\gamma^{j-i}\tilde r_j\mid x_i,c_i=c\bigr]$,
where $\pi$ governs subsequent family choices under the fixed coding agent and API library. The value $Q_\theta(x_i,c)$ estimates the remaining skill- and task-level return conditional on choosing family $c$ next.
After collecting $\mathcal{D}_{\mathrm{SCB}}$ across the training tasks, we optimize $Q_\theta$ on this fixed transition dataset using Bellman targets rather than directly regressing the rollout returns $G_i$ or tree estimates $\widehat Q_i(c)$. All evaluated tree edges, including failed and nonpreferred branches, contribute to this update.
For transitions sampled from $\mathcal{D}_{\mathrm{SCB}}$, we optimize $\theta$ by gradient-based backpropagation through $Q_\theta$, minimizing the following Smooth-$L_1$ Bellman loss:
\begin{equation}
\label{eq:eal-loss}
\mathcal{L}_{\mathrm{EAL}}(\theta)
=
\mathbb{E}_{(x_i,c)\sim\mathcal{D}_{\mathrm{SCB}}}
\ell\left[
Q_\theta(x_i,c)
-
\bigl(
\tilde r_i^c+\gamma(1-d_i^c)\max_{c'\in\mathcal{C}_{i+1}}Q_{\bar\theta}(x_{i+1}^c,c')
\bigr)
\right],
\end{equation}
where $Q_{\bar\theta}$ is a stop-gradient target copy and $\ell(\epsilon)$ refers to the Smooth-$L_1$ loss, which maintains stable gradients in the regime of large errors as the $L_1$ loss does and facilitates smooth optimization for small errors analogous to the $L_2$ loss:
\begin{equation}
\label{eq:smooth-l1}
\ell(\epsilon)
=
\begin{cases}
\tfrac{1}{2}\epsilon^{2} & \text{if }|\epsilon|<1,\\
|\epsilon|-\tfrac{1}{2} & \text{otherwise}.
\end{cases}
\end{equation}

\textbf{Q-Modeling}.
$Q_\theta(x_i,c)$ is parameterized with a lightweight VLM backbone (\textit{e.g.} Qwen3.5-0.8B).
The state $x_i=(\ell,o_i,h_i,z_i,g_i)$ and family $c$ form a multimodal token sequence ending in a \texttt{<VALUE>} token. An MLP head maps its final-layer hidden state to the scalar Q-value, optimized with the Bellman objective in~\cref{eq:eal-loss}.

\subsubsection{Test-Time Skill Coordination}

During training, repeated access to $\xi_i$ and oracle outcome judgments supervises values, whereas neither is included at test time.
The learned values induce a relative responsibility preference over admissible families:
\begin{equation}
\label{eq:responsibility-preference}
\rho_\theta(c\mid x_i)=\frac{\exp\!\left(Q_\theta(x_i,c)\right)}{\sum_{c'\in\mathcal C_i}\exp\!\left(Q_\theta(x_i,c')\right)}.
\end{equation}
The \algo{} coordinator selects $c_i^*=\arg\max_{c\in\mathcal C_i}\rho_\theta(c\mid x_i)$ (equivalently $\arg\max_c Q_\theta(x_i,c)$; see~\Cref{app:responsibility-preference}), and the coding agent generates $a_i\sim\codepolicy(\cdot\mid x_i,c_i^*)$.
The coding agent executes the selected block until the next semantic boundary or task termination, then updates $(h_i,z_i,g_i)$ from observable feedback.
Family selection repeats at each subsequent boundary with $Q_\theta$ fixed.
Both family selection and local code generation respond to updated observations without parameter updates or changes to executor interfaces.

\section{Experiments}
\label{sec:experiments}

\subsection{Experimental Protocol}
\textbf{Benchmarks}.
Our evaluation contains 100 tabletop-manipulation tasks across three benchmarks: LIBERO-Pro~\citep{LIBERO-Pro}, RoboSuite~\citep{RoboSuite}, and RoboTwin~2.0~\citep{RoboTwin}.
We use scene seed~0 only to collect privileged \scb{} branches and train $Q_\theta$.
The learned coordinator is then frozen and evaluated on seeds 1--10, giving 1,000 test episodes in total.


\textbf{Baselines.} We compare \algo{} with the following baselines:
\begin{itemize}[leftmargin=*]

  \item \textit{End-to-end policies}:
  $\pi_0$~\citep{pi0}, $\pi_{0.5}$~\citep{pi05}, FastWAM~\citep{FastWAM}, FasterWAM~\citep{FasterWAM}, as standalone controllers;

  \item \textit{Code-as-policy}:
  CaP~\citep{CaP}, CaP-X~\citep{Cap-X}, RATS~\citep{RATS}, and ASPIRE~\citep{ASPIRE}, without learned family coordination.

  \item \textit{VLA with harness}:
  Pigey~\citep{Pigey}, VLS~\citep{VLS}, and Harness-VLA~\citep{HarnessVLA}, which invoke a frozen policy.
\end{itemize}

\subsection{Benchmark Performance}
\label{sec:benchmark-performance}
\begin{table}[htbp]
  \centering
  \caption{Aggregate success rates (\%) on LIBERO-Pro (instruction-redirection \textbf{T} / position-swap \textbf{S}), RoboSuite, and RoboTwin.
RoboSuite and RoboTwin use their full task pools under seeds $1$--$10$.
Overall is the mean over available columns.
On RoboTwin, $\pi_0$ and $\pi_{0.5}$ are incompatible with RoboTwin's dual-arm setup, so results involving these models are marked ``/'' because of the embodiment mismatch.
For fair comparison, global cross-episode memory is disabled for baselines that rely on memory.
All seeds are tested in \textbf{one trial}.
}
  \label{tab:main_exp}
  \small
  \setlength{\tabcolsep}{2.4pt}
  \resizebox{\linewidth}{!}{%
  \begingroup
  \setlength{\aboverulesep}{0pt}%
  \setlength{\belowrulesep}{0pt}%
  \renewcommand{\arraystretch}{1.14}%
  \newcommand{\ov}[1]{\multicolumn{1}{>{\columncolor{gray!20}[\tabcolsep][0pt]}c@{}}{#1}}%
  \begin{tabular}{@{}l cccccccc cc c@{}}
  \toprule
  \addlinespace[0.0001pt]
  \multirow{2}{*}{Method} & \multicolumn{8}{c}{LIBERO-Pro} & \multirow{2}{*}{RoboSuite} & \multirow{2}{*}{RoboTwin} & \multirow{2}{*}{Overall} \\
  \cmidrule(lr){2-9}
  \addlinespace[0.0001pt]
  & Spat-T & Spat-S & Obj-T & Obj-S & Goal-T & Goal-S & L10-T & L10-S & & & \\
  \midrule
  \addlinespace[0.0001pt]
  \multicolumn{11}{l}{\textit{End-to-end policies}} & \ov{} \\
  $\pi_0$~\citep{pi0}                      & 0.0 & 0.0 & 0.0 & 2.0 & 0.0 & 0.0 & 0.0 & 0.0 &  0.0 &  / & \ov{0.2} \\
  $\pi_{0.5}$~\citep{pi05}                              & 1.0 & 20.0 & 1.0 & 17.0 & 2.0 & 38.0 & 1.0 & 8.0 &  0.0 &  / & \ov{9.8} \\
  FastWAM~\citep{FastWAM}                                  & 30.0 & 0.0 & 10.0 & 0.0 & 10.0 & 0.0 & 10.0 & 0.0 & 0.0 & 81.9 & \ov{14.2} \\
  FasterWAM~\citep{FasterWAM}                                & 50.0 & 0.0 & 10.0 & 10.0 & 10.0 & 10.0 & 10.0 & 0.0 & 0.0 & 85.6 & \ov{18.6} \\
  \midrule
  \addlinespace[0.0001pt]
  \multicolumn{11}{l}{\textit{Code as policy}} & \ov{} \\
CaP~\citep{CaP}                          & 0.0 & 0.0 & 0.0 & 0.0 & 0.0 & 0.0 & 0.0 & 0.0 & 70.0 & 5.6 & \ov{7.6} \\
CaP-X~\citep{Cap-X}                      & 0.0 & 0.0 & 10.0 & 0.0 & 0.0 & 0.0 & 0.0 & 0.0 & 50.0 & 6.9 & \ov{6.7} \\
RATS~\citep{RATS}                        & 2.0 & 18.0 & 10.0 & 8.0 & 8.0 & 8.0 & 0.0 & 0.0 & 45.0 & 0.0 & \ov{9.9} \\
  ASPIRE~\citep{ASPIRE}          & 60.0 & 51.0 & \textbf{95.0} & \textbf{98.0} & 45.0 & 81.0 & 38.0 & 22.0 & 80.0 & 0.0 & \ov{57.0} \\
  \midrule
  \addlinespace[0.0001pt]
  \multicolumn{11}{l}{\textit{VLA with harness}} & \ov{} \\
  Pigey~\citep{Pigey}
& \textbf{80.0} & 66.0 & 54.0 & 54.0 & 22.0 & 44.0 & 28.0 & 12.0 & 15.0 & / & \ov{41.7} \\
  VLS~\citep{VLS}
& 54.0 & 42.0 & 41.0 & 45.0 & 33.0 & 38.0 & 25.0 & 15.0 & 0.0 & / & \ov{32.6} \\
  Harness-VLA~\citep{HarnessVLA}           & 50.0 & \textbf{70.0} & 50.0 & 90.0 & 40.0 & 60.0 & 40.0 & 30.0 & 20.0 & / & \ov{50.0} \\
  \textbf{\algo{} (ours)} & 68.0 & 68.0 & 92.0 & 86.0 & \textbf{66.0} & \textbf{84.0} & \textbf{64.0} & \textbf{62.0} & \textbf{90.0} & \textbf{90.0} & \ov{\textbf{77.0}} \\
  \bottomrule
  \end{tabular}%
  \endgroup
  }
\end{table}

\textbf{End-to-end policies struggle with whole-task distribution shifts.}
For both VLAs and WAMs, unfamiliar skill combinations, scene types, and perturbations may deviate substantially from their training distributions, leading to low zero-shot SRs in \Cref{tab:main_exp}.
In comparison, \algo{} generates composable modular alternatives under such conditions while retaining end-to-end policies for local skills suited to their capabilities.

\textbf{Code-as-policy methods are constrained by contact precision and accumulated experience.}
Code-as-policy methods offer stronger long-term planning and composability for long-horizon bimanual tasks such as those in RoboTwin.
However, within-skill contact-rich performance depends on available primitives: CaP and CaP-X generate detailed code at test time, yet may lack the precision needed for tasks such as placing a mug inside a microwave in LIBERO-Pro (see \Cref{fig:ablation-hard}).
Single-turn evaluation without cross-episode adaptation or multi-turn memory further limits the benefits of accumulated experience for memory-based methods like CaP-X, RATS, and ASPIRE.
\algo{} addresses these limitations by combining modular composition with local end-to-end control and learning state-dependent family responsibility from training-time outcomes.

\textbf{VLA harnesses remain limited by outcome anticipation.}
We observe that VLS~\citep{VLS} can be misled by early-stage behavior in long-horizon tasks, where initial progress may not reflect readiness for later interactions.
Harness-VLA is closely related to our work and achieves competitive performance. 
However, once the selected family changes the physical state, the alternative family's outcome at the original state remains unobserved, so family switching relies on experiential judgments.
\algo{} addresses this gap through \scb{} and \eal{}, which expose same-state alternatives for family selection.

\subsection{Experimental Analysis}
\label{sec:experimental-analysis}

\begin{figure}[!htbp]
  \centering
  \def\abImgH{3.2cm}%
  \begin{minipage}[t]{\dimexpr \abImgH*670/373\relax}
    \centering
    \includegraphics[height=\abImgH]{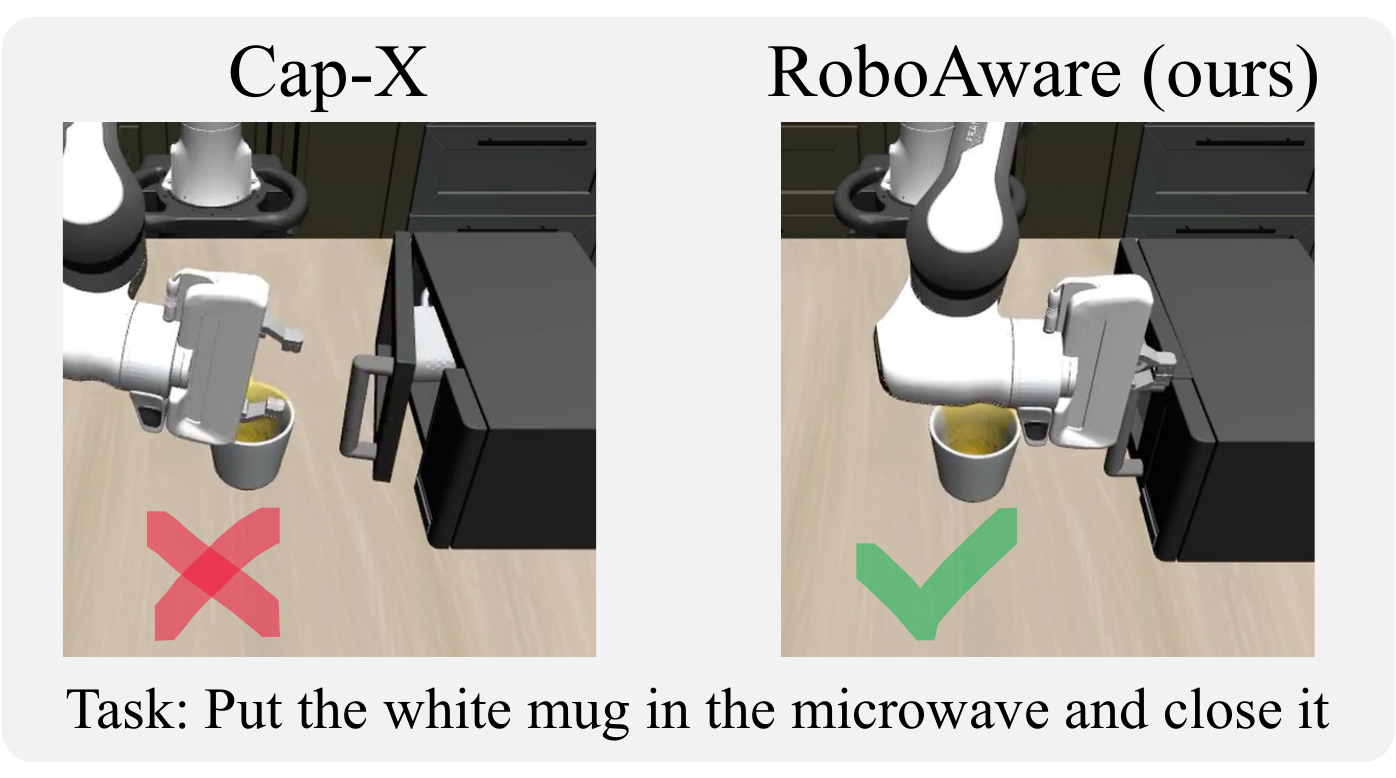}%
    \captionof{figure}{Comparison between CaP-X and \algo{} on a contact-rich LIBERO-Pro task ``Put the white mug in the microwave and close it.''}
    \label{fig:ablation-hard}
  \end{minipage}%
  \hspace{0.02\linewidth}%
  \begin{minipage}[t]{\dimexpr \abImgH*897/374\relax}
    \centering
    \includegraphics[height=\abImgH]{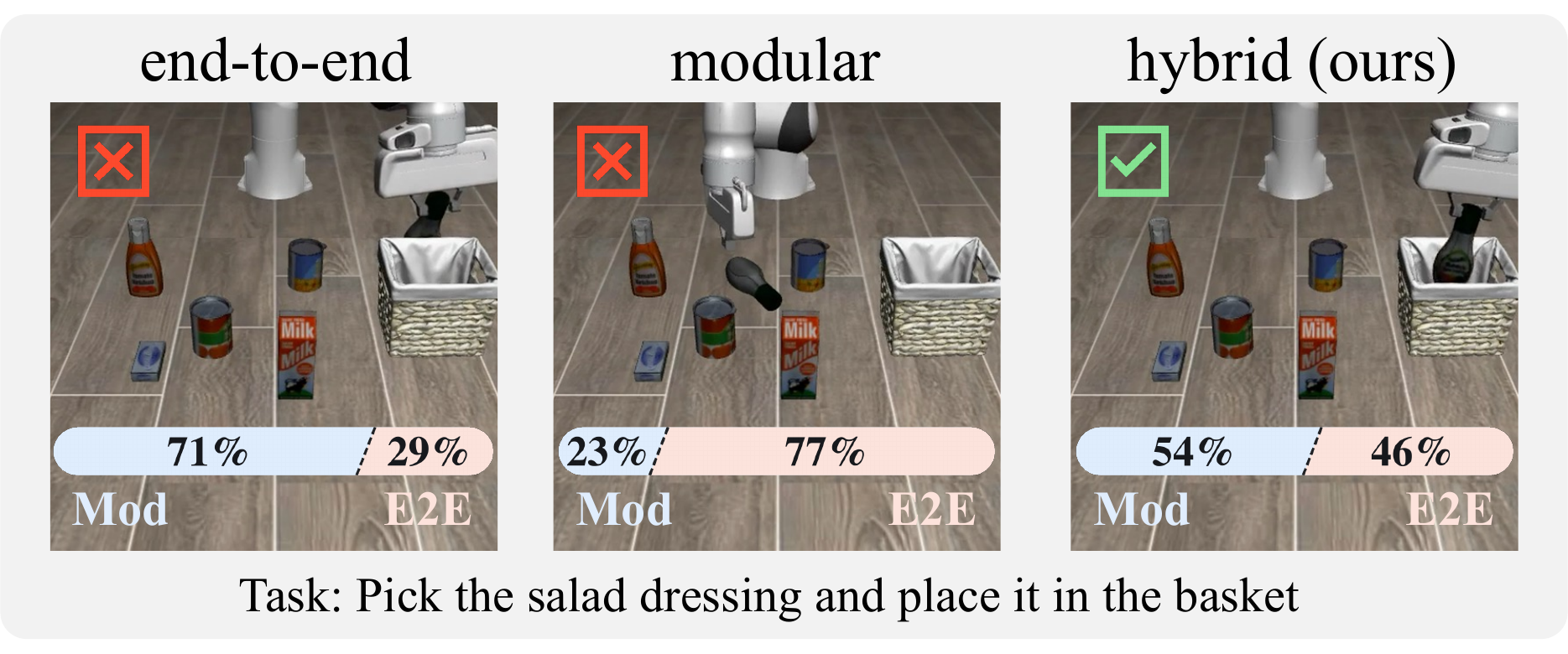}%
    \captionof{figure}{A representative LIBERO-Pro task comparison and visualization of the responsibility preference distribution $\rho_\theta$ for \algo{}, \algo{}-EO, and \algo{}-MO.}
    \label{fig:ablation-hybrid}
  \end{minipage}
\end{figure}

\textbf{Takeaway~1: Neither policy family alone provides broad high-success task coverage.}
We restrict code generation to modular composition (\algo{}-MO) or end-to-end blocks (\algo{}-EO), keeping the same API library and tasks as \algo{}.
Success rate (SR) reductions in group~A of \Cref{tab:ablation-analysis} show that both families must be coordinated by state.
In a LIBERO-Pro task illustrated in \Cref{fig:ablation-hybrid}, \algo{}-EO and \algo{}-MO fail at different stages, whereas \algo{} succeeds by switching policy families across \pschema{} boundaries.
The $\rho_\theta$ visualization assigns lower preference to the families used by the ablation variants at the illustrated states, consistent with their failures.

\begin{table}[H]
  \begin{minipage}[t]{0.62\linewidth}
    \centering
    \captionsetup{justification=raggedright,singlelinecheck=false}
    \caption{Ablation success rates (\%) on the three benchmarks, grouped by conclusions~A--C in this analysis. \algo{} is the common reference. LIBERO-Pro aggregates its eight T/S evaluation splits. Abbreviations: MO\,=\,modular only; EO\,=\,end-to-end only; UO\,=\,unpaired outcomes; TF\,=\,training free; FH\,=\,fixed handoff; ES\,=\,episode selection.}
    \label{tab:ablation-analysis}
    \small
    \setlength{\tabcolsep}{5pt}
    \begin{tabular}{@{}lccc@{}}
      \toprule
      Variants & LIBERO-Pro & RoboSuite & RoboTwin~2.0 \\
      \midrule
      \multicolumn{4}{@{}l}{\textit{A. Policy Family}} \\
      \algo{}-MO & 54.8 & 67.5 & 41.9 \\
      \algo{}-EO & 49.5 & 60.0 & 45.6 \\
      \midrule
      \multicolumn{4}{@{}l}{\textit{B. Counterfactual Outcome}} \\
      \algo{}-UO & 58.5 & 75.0 & 53.8 \\
      \algo{}-TF & 51.0 & 60.0 & 40.0 \\
      \midrule
      \multicolumn{4}{@{}l}{\textit{C. Coordination Mode}} \\
      \algo{}-FH & 64.5 & 82.5 & 55.6 \\
      \algo{}-ES & 59.3 & 70.0 & 51.9 \\
      \midrule
      \algo{} & \textbf{73.8} & \textbf{90.0} & \textbf{90.0} \\
      \bottomrule
    \end{tabular}
  \end{minipage}%
  \hfill
  \begin{minipage}[t]{0.36\linewidth}
    \centering
    \captionsetup{justification=raggedright,singlelinecheck=false}
    \caption{Group~B ablations of coordinator training and same-state outcomes. \ding{51}: present; \ding{55}: absent.}
    \label{tab:paired-supervision}
    \small
    \setlength{\tabcolsep}{4pt}
    \begin{tabular}{@{}lcc@{}}
      \toprule
      Variant & Training & Counterfactual \\
      \midrule
      \textit{ours} & \ding{51} & \ding{51} \\
      \textit{ours}-UO & \ding{51} & \ding{55} \\
      \textit{ours}-TF & \ding{55} & \ding{55} \\
      \bottomrule
    \end{tabular}
    \vspace{1.6em}
    \captionsetup{position=top,justification=raggedright,singlelinecheck=false,skip=1em}
    \captionof{table}{Group~C ablations of when family responsibility may change. When: decision points; Family: modular / e2e.}
    \label{tab:coordination-strategies}
    \begin{tabular}{@{}p{0.28\linewidth}@{}p{0.4\linewidth}@{}p{0.32\linewidth}@{}}
      \toprule
      Variant & When & Family \\
      \midrule
      \textit{ours} & Each stage & Mod / E2E \\
      \textit{ours}-FH & Fixed handoff & Mod + E2E  \\
      \textit{ours}-ES & Episode start & Mod / E2E \\
      \bottomrule
    \end{tabular}
  \end{minipage}
\end{table}

\textbf{Takeaway~2: Paired same-state outcomes improve Execution-aware learning.}
To evaluate the importance of \scb{}, we ablate coordinator training and counterfactual supervision (\Cref{tab:paired-supervision}): \algo{} uses both, \algo{}-UO trains $Q_\theta$ only on executed trajectories without state-locked branching, and \algo{}-TF skips coordinator training so the coding agent chooses directly from the hybrid API library.
Higher SR for \algo{}-UO than \algo{}-TF in group~B of \Cref{tab:ablation-analysis} indicates that the coding agent lacks inherent execution awareness and that coordinator training still helps without counterfactuals.
Yet executed-only branches inherit the agent's experience distribution and introduce selection bias.
A modest SR decrease for \algo{}-UO relative to \algo{} therefore supports the necessity of paired counterfactual executions from the same $\xi_i$.

\textbf{Takeaway~3: \algo{} benefits from fine-grained coordination at \pschema{} stages.}
Following the same training protocol to obtain $Q_\theta$, we compare three coordination strategies that differ in when family responsibility may change.
\algo{} reselects a policy family at every \pschema{} stage.
\algo{}-FH (fixed handoff) follows a fixed protocol: modular APIs for \textit{perceive} and \textit{pre-manipulate}, then an \eetfamily{} call for subsequent execution.
\algo{}-ES (episode selection) chooses one family at episode start and retains it for the entire task.
Lower SRs for both variants in group~C of \Cref{tab:ablation-analysis} suggests that neither a fixed handoff nor a single episode-level decision tracking changing execution responsibility as well as per-stage coordination (\Cref{tab:coordination-strategies}).

\begin{figure}[H]
  \centering
  \begin{minipage}[t]{0.5\linewidth}
    \vspace{0pt}
    \centering
    \includegraphics[width=\linewidth]{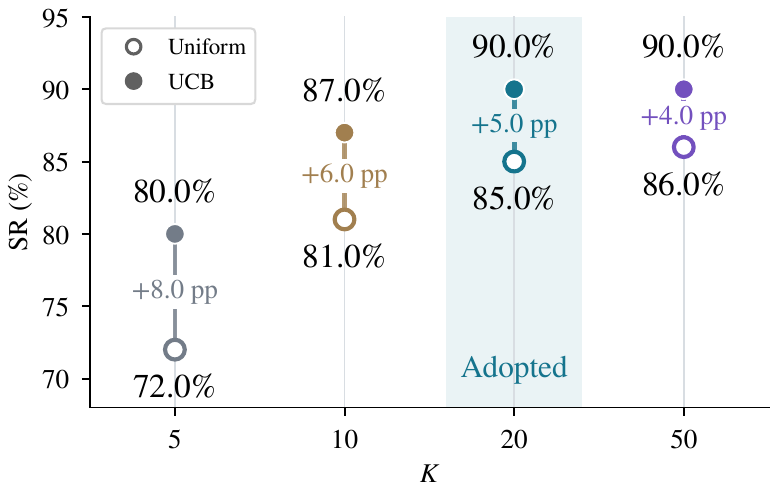}
    \captionof{figure}{Average RoboSuite success rates (\%) across budgets $K\in\{5,10,20,50\}$. Hollow \& filled circles denote \algo{}-UA \& \algo{}.
    We adopt $K{=}20$.}
    \label{fig:explore-budget}
  \end{minipage}\hfill
  \begin{minipage}[t]{0.46\linewidth}
    \vspace{0pt}
    \centering
    \includegraphics[width=\linewidth]{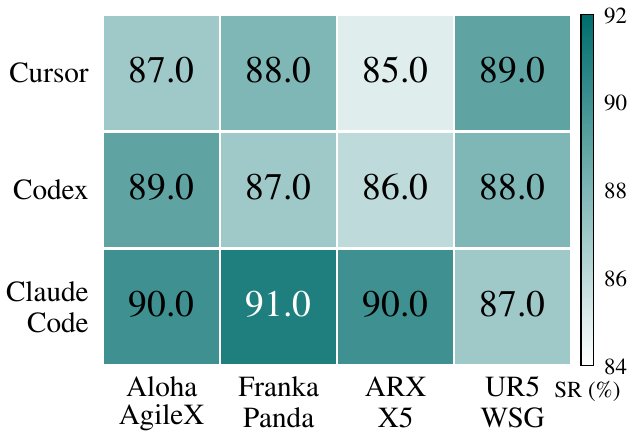}
    \captionof{figure}{RoboTwin~2.0 success rates (\%) on a four-task subset across coding-agent backends (rows) and embodiments (columns). Color bar shows the SR range.}
    \label{fig:agent-backend}
  \end{minipage}
\end{figure}

\textbf{Takeaway~4: UCB-guided exploration uses a limited branching budget more effectively.}
To test how a finite simulation budget is allocated during branching, we compare \algo{} with \algo{}-UA, which allocates evaluations uniformly over admissible families instead of the UCB rule in~\cref{eq:ucb}.
\Cref{fig:explore-budget} reports both under $K\in\{5,10,20,50\}$ in RoboSuite. SR gains diminish with $K$, so we use the elbow $K{=}20$.
Under matched budgets, \algo{} exceeds \algo{}-UA, indicating that UCB concentrates evaluations on promising families more effectively than uniform allocation.

\textbf{Takeaway~5: \algo{} is less sensitive to coding-agent and embodiment variations.}
To assess sensitivity to changes in the coding agent and robot embodiment, we evaluate \algo{} with three coding-agent backends (Claude Code as default, Cursor, and Codex) across four embodiments (Aloha-AgileX as default, Franka-Panda, ARX-X5, and UR5-WSG) on four tasks in RoboTwin 2.0 (See Appendix \Cref{app:benchmark-details} for details), retaining the same API interfaces.
\Cref{fig:agent-backend} shows a small average SR gap across coding-agent backends and embodiments.
We attribute the small backend gap to \pschema{}, which limits each generation to one code block $a_i\in\mathcal{A}_{c_i}(x_i,z_i,g_i)$.
Notably, the cuRoboV2 backend of \texttt{Plan\_to\_eef\_pose} inherently supports different robots through the same API interface across embodiments.


\section{Related Work}
\subsection{Embodied Coding Agents}
Embodied coding agents have progressed along three lines. Early language planners translated goals into sequences over closed skill APIs and improved grounding through semantic matching, affordance scoring, program-structured prompting, and execution feedback~\citep{LLM-Planner,SayCan,ProgPrompt,InnerMonologue}. Executable-policy methods such as CaP~\citep{CaP}, VoxPoser~\citep{VoxPoser}, RoboCodeX~\citep{RoboCodeX}, and CodeDiffuser~\citep{CodeDiffuser} then generated Python controllers or geometric intermediates, gaining compositionality, spatial reasoning, and interpretability while remaining bounded by designer-provided abstractions and API coverage. Recent systems make the program itself adaptive. CaP-X~\citep{Cap-X} and AOR~\citep{AOR} repair code from multimodal rollouts, RHO~\citep{RHO} and GaP~\citep{GaP} optimize repository- or graph-structured policies through execution, VLCP~\citep{VLCP} rewrites failed controllers within an episode, PhysCaP~\citep{PhysCaP} actively probes latent physical properties, and ASPIRE~\citep{ASPIRE} compounds validated repairs into a reusable skill library. This line strengthens modular program execution, yet its evidence is conditional on generating a modular code block. Neither success nor code-only repair reveals whether an end-to-end block performs better at the same state.
We instead use \scb{} to generate and execute modular and end-to-end coding blocks from the same restored state, and \eal{} learns their family-conditioned values from paired outcomes.

\subsection{End-to-End Policy Orchestration and Execution Harnesses}
End-to-end policy orchestration has likewise progressed along three lines: generalist VLAs including RT-1~\citep{RT-1}, RT-2~\citep{RT-2}, Octo~\citep{Octo}, OpenVLA~\citep{OpenVLA}, and $\pi_0$~\citep{pi0} broaden local language-conditioned control, while hierarchical agents move planning, memory, grounding, and progress tracking to a VLM or code layer, exposing VLAs as bounded tools.
Recent variants add tool-aligned post-training, persistent semantic state, tractability-aware interfaces, interruptible physical orchestration, and modular monitor, planner, and controller wrappers~\citep{HVLA,TaLVLA,Pigey,HarnessVLA,CodeGraphVLP,Cortex,VoLo,RoboBRIDGE}. VLS~\citep{VLS} instead steers the sampling of a frozen generative policy with VLM-synthesized rewards. Hybrid and heterogeneous-policy harnesses including RoboRouter~\citep{RoboRouter}, RoboHarness~\citep{RoboHarness}, and RouterVLA~\citep{RouterVLA} then infer applicability from execution traces, semantic retrieval, capability memory, outcome-disjoint probes, or transition structure.
Related methods including Reset~\citep{Reset}, SwitchVLA~\citep{SwitchVLA}, SemHandoff~\citep{SemHandoff}, and BICPO-VLA~\citep{BICPO-VLA} prepare out-of-distribution initial states, condition switching on execution phase, diagnose semantic handoffs, or reduce asynchronous takeover mismatch.
Finally, VLM oracles, learned rewards, frozen-feature success probes, counterfactual relabeling, and self-evolving critics detect, rank, or repair attempts~\citep{ARMOR,VISOR,TOPReward,LRM,RoboMonkey,Dream2Act,CAST,VLAProbe,CounterAlign,Zetta}.
These methods estimate absolute success, compare within-policy alternatives, or infer cross-policy suitability from retrospective traces and probes. 
Our distinction is to use simulator-restored, same-state executions to obtain paired outcomes for modular and end-to-end families within a common skill-stage interface.

\section{Conclusion and Limitations}
\label{sec:conclusion-limitations}
We presented \algo{}, which learns state-dependent responsibility between modular composition and end-to-end control while keeping the coding agent and API library fixed.
The \pschema{} skill schema defines the semantic boundaries at which responsibility is assigned, and \scb{} exposes both family outcomes at such a boundary by executing paired code blocks from identical restored states.
From this paired experience, \eal{} combines UCB-guided MCTS with Q-learning into family-conditioned estimates of remaining skill- and task-level return.
At deployment, these estimates select a family before code generation from observable context alone, and responsibility is reconsidered at each subsequent boundary without simulator restoration, privileged inputs, or online search.
Across 100 tasks in LIBERO-Pro, RoboSuite, and RoboTwin~2.0, this makes anticipated physical outcomes an explicit basis for composing complementary frozen skills.

We outline two directions for future work.
First, \algo{} currently integrates only modular composition and end-to-end
policies. Extending it to generalizable reinforcement learning policies,
retrieval-based policies, and policies learned from demonstrations may require
new counterfactual outcome data and further coordinator training.
Second, responsibility also changes hands only at \pschema{} boundaries, so a single
atomic end-to-end invocation may span several stages and leave transient
failures to its internal controller until a postcondition, failure signal, or
timeout returns control. Finer-grained interruption under ambiguous
observations therefore remains an open problem.

\phantomsection{}
\addcontentsline{toc}{section}{References}
\begingroup
\setlength{\bibsep}{0.2pt plus .2pt minus .1pt}
\bibliographystyle{unsrtnat}
\bibliography{references}
\endgroup

\setcounter{topnumber}{2}
\setcounter{bottomnumber}{2}
\setcounter{totalnumber}{4}
\renewcommand{\bottomfraction}{0.8}
\clearpage
\appendix
\setcounter{figure}{0}
\setcounter{table}{0}
\renewcommand{\theHfigure}{appendix.\arabic{figure}}
\renewcommand{\theHtable}{appendix.\arabic{table}}
\newcommand{\mainCref}[1]{\Cref{#1} in the main text}

\addtocontents{toc}{\protect\setcounter{tocdepth}{3}}
\begingroup
  \renewcommand{\contentsname}{Appendix Contents}
  \setcounter{tocdepth}{-1}
  \hypersetup{linkcolor=black}
  \fontsize{9}{10}\selectfont
  \setlength{\parskip}{0pt}
  \tableofcontents
\endgroup
\clearpage

\makeatletter
\setlength{\@fptop}{0pt}
\setlength{\@fpsep}{14pt}
\makeatother

\section{Methodological Details}\label{app:method-details}

\subsection{\texorpdfstring{\pschema{}}{P5} Stage Contracts}\label{app:stage-contracts}
Each \pschema{} stage declares an entry predicate, the typed inputs it requires, an admissible-family set $\mathcal{C}_i(z_i,g_i)$, the APIs a family-constrained code block may call, a success postcondition, and the failure or timeout feedback that chooses the successor stage.
\Cref{tab:stage-contracts} summarizes these contracts; typed API I/O appears in \Cref{tab:embodied-apis-io}.
The coordinator selects a family only when the coding agent has control and the current stage's entry predicate holds.
Once an atomic API call begins, that selection remains fixed until the call returns after satisfying its registered postcondition, reporting a failure, or exhausting its step budget.

\begin{table}[H]
\centering
\caption{
\pschema{} stage contracts and admissible families.
At stage $z_i$ of skill $g_i$, the coordinator selects $c_i\in\mathcal{C}_i(z_i,g_i)$ before code generation; APIs must match $(z_i,c_i)$ (\mainCref{tab:embodied-apis}).
Columns: $\phi^{\mathrm{in}}$---entry predicate; $\mathcal{C}_i$---admissible families; $\phi^{\mathrm{out}}$---registered success postcondition; Fail---successor after failure/timeout (retry $z_i$, return upstream $\uparrow$, or terminate $\bot$).
Under $\modfamily$, $\phi^{\mathrm{out}}$ is the typed modular return (e.g.\ pose at \textit{perceive}, eef pose at \textit{propose}); under $\eetfamily$, it is covered by the atomic call postcondition.
\textit{propose} is Mod-only; an E2E call may complete the skill without visiting it.
}
\label{tab:stage-contracts}
\footnotesize
\setlength{\tabcolsep}{3.5pt}
\begin{tabular}{@{}p{0.14\linewidth}p{0.24\linewidth}p{0.12\linewidth}p{0.22\linewidth}p{0.18\linewidth}@{}}
\toprule
$z_i$ & $\phi^{\mathrm{in}}$ & $\mathcal{C}_i$ & $\phi^{\mathrm{out}}$ & Fail \\
\midrule
\textit{perceive}
  & target ID/pose $\notin h_i$
  & $\{\modfamily,\eetfamily\}$
  & typed target info
  & retry / $\bot$ \\
\textit{propose}
  & pose/mask $\in h_i$ $\land$ eef target $\notin h_i$
  & $\{\modfamily\}$
  & valid eef pose
  & retry / $\uparrow$\textit{perceive} / $\bot$ \\
\textit{pre-manipulate}
  & approach target available
  & $\{\modfamily,\eetfamily\}$
  & pre-contact reached
  & replan / $\uparrow$ / $\bot$ \\
\textit{perform}
  & pre-manip.\ $\phi^{\mathrm{out}}$
  & $\{\modfamily,\eetfamily\}$
  & contact/gripper postcond.
  & retry / $\rightarrow$\textit{post-manip.} / $\bot$ \\
\textit{post-manipulate}
  & perform $\phi^{\mathrm{out}}$
  & $\{\modfamily,\eetfamily\}$
  & retreat/release/stabilize
  & retry / $\uparrow$ / $\bot$ \\
\bottomrule
\end{tabular}
\end{table}

\paragraph{Required inputs and permitted APIs.}
Under $\modfamily$, the permitted API depends on the current stage:
\texttt{Perceive\_object\_pose} at \textit{perceive},
\texttt{Propose\_eef\_pose} at \textit{propose},
\texttt{Plan\_to\_eef\_pose} at \textit{pre-manipulate} and \textit{post-manipulate},
and \texttt{Perform\_gripper} at \textit{perform}.
A modular code block may compose only APIs admissible for the current $z_i$ (and any immediately chained modular helpers that do not cross a new responsibility boundary).
Under $\eetfamily$, the sole permitted pattern is \performendtoend{}(\texttt{route\_id},\ldots), whose declared coverage is \textit{perceive}, \textit{pre-manipulate}, \textit{perform}, and \textit{post-manipulate} (\mainCref{tab:embodied-apis}).
After the coordinator selects $\eetfamily$, the coding agent supplies \texttt{route\_id} and the required typed observations in $a_i$. Thus, $Q_\theta$ selects the family, while the coding agent selects the backend within that family.

\paragraph{Why \textit{propose} admits only $\modfamily$.}
\texttt{Propose\_eef\_pose} is the only API assigned to \textit{propose}, where it produces the end-effector target required by modular motion planning.
End-to-end backends determine grasp and approach targets internally and do not expose them through this interface, so \performendtoend{} does not cover \textit{propose}.
Consequently $\mathcal{C}_i=\{\modfamily\}$ whenever $z_i$ is \textit{propose}, so no E2E comparison is issued at that boundary.
A skill enters \textit{propose} only when it requires an explicit end-effector target. An E2E call beginning at \textit{perceive} or a later admissible stage can therefore complete the skill without entering \textit{propose}.

\paragraph{Multi-stage atomic returns and stage advancement.}
A single \performendtoend{} call may satisfy several consecutive stage postconditions before returning, but its family assignment remains fixed throughout the call.
When control returns, the coding agent updates $(h_i,z_i,g_i)$ from observable feedback only: typed returns, registered postcondition bits, failure codes, and timeout flags.
The next decision stage $z_{i+1}$ is the earliest stage in the active skill whose entry predicate still holds and whose success postcondition is unmet.
Stages whose postconditions have already been satisfied are skipped, so the coordinator makes no additional family selections for them unless execution later reaches an entry where those postconditions are unmet.
If the call fails or times out after partial progress, $z_{i+1}$ is likewise the earliest unmet stage that remains enterable under the returned feedback (often the stage at which the postcondition failed), enabling retry or upstream fallback as in \Cref{tab:stage-contracts}.
If every stage postcondition of $g_i$ is met, the skill completes and the plan advances to the next schema in $\mathcal{G}$.
If execution must continue but $\mathcal{C}_{i+1}$ is empty, the episode terminates with failure, as specified in \Cref{alg:roboaware-deploy}.

\subsection{\algo{}: Training and Deployment}\label{app:deployment-algo}

\begin{algorithm}[H]
  \caption{\algo{}: MCTS data collection, Q-learning, and deployment}\label{alg:roboaware-deploy}
  \SetAlgoLined{}
  \SetAlgoNoEnd{}
  \SetAlFnt{\small}
  \SetAlgoNlRelativeSize{-1}
  \LinesNumbered{}
  \KwIn{training tasks $\mathcal{T}_{\mathrm{train}}$; test instruction $\ell_{\mathrm{test}}$; simulator $\mathcal E$; $\codepolicy$ and API library; budget $K$}
  \KwOut{trained coordinator $Q_\theta$; test execution history $h$ and terminal reward $r$}
  Freeze $\codepolicy$ and the API library; initialize $\theta$, $\bar\theta\leftarrow\theta$, and $\mathcal D_{\mathrm{SCB}}\leftarrow\emptyset$\;
  \tcp{Phase 1. Offline training: prepare search roots}
  Collect roots $\mathcal R$ from training-task \pschema{} executions using the neighboring-state sampling protocol described below, with matching histories\;
  \ForEach{sampled root $\xi_t\in\mathcal R$}{
    Initialize tree $\mathcal T$ at $\xi_t$ and remaining budget $b\leftarrow K$\;
    \While{$b>0$}{
      \tcp{1. Selection}
      Follow stored paths by UCB (\cref{eq:ucb}) to a nonterminal node $j$; select a new node only if $b\geq|\mathcal C_j|$, otherwise reevaluate an expanded node\;
      \tcp{2. Expansion}
      \eIf{node $j$ is new}{
        Set $N_j=N_j(c)=0$, $\widehat Q_j(c)=Q_\theta(x_j,c)$, and $A_j\leftarrow\mathcal C_j$\;
      }{
        $A_j\leftarrow\{c_j^{\mathrm{tree}}\}$ by \cref{eq:ucb}\;
      }
      \ForEach{$c\in A_j$}{
        Restore $\xi_j$ and its history; generate $a_j^c\sim\codepolicy(\cdot\mid x_j,c)$ and execute to the next boundary or termination\;
        Record $(x_{j+1}^c,\xi_{j+1}^c,d_j^c)$ and compute $\tilde r_j^c$ from the skill/task oracle scores\;
        Attach the successor to $\mathcal T$; append $(x_j,c,\tilde r_j^c,x_{j+1}^c,d_j^c,\xi_{j+1}^c)$ to $\mathcal D_{\mathrm{SCB}}$\;
        \tcp{3. Rollout}
        Continue a nonterminal successor to a stopping condition without tree expansion; obtain $G_{j+1}$, or set $G_{j+1}=0$ if $d_j^c=1$\;
        $G_j\leftarrow\tilde r_j^c+\gamma G_{j+1}$\;
        \tcp{4. Backpropagation}
        Back up $G_j$ and prefix rewards along the selected path; increment visits, average returns into $\widehat Q$, and set $b\leftarrow b-1$\;
      }
    }
  }
  \tcp{Q-learning after data collection}
  Fit $Q_\theta$ on all of $\mathcal D_{\mathrm{SCB}}$ for $M$ minibatch updates using \cref{eq:eal-loss,eq:smooth-l1}, learning rate $\eta$, and target refresh interval $J$\;
  \tcp{Phase 2. Deployment with fixed parameters}
  Observe $o_0$; generate test plan $\mathcal G$ from $(\ell_{\mathrm{test}},o_0)$; initialize $h_0\leftarrow\emptyset$, $(z_0,g_0)$, and $i\leftarrow0$\;
  \While{the task is nonterminal and the episode horizon remains}{
    Build $x_i=(\ell_{\mathrm{test}},o_i,h_i,z_i,g_i)$ and admissible families $\mathcal C_i(z_i,g_i)$\;
    \lIf{$\mathcal C_i=\emptyset$}{Terminate with failure and exit the loop}
    $c_i^\star\leftarrow\arg\max_{c\in\mathcal C_i}\rho_\theta(c\mid x_i)$; generate $a_i\sim\codepolicy(\cdot\mid x_i,c_i^\star)$\;
    Execute $a_i$ to the next boundary or termination, respecting atomic APIs\;
    Observe feedback; update $x_{i+1}$ and $\mathcal G$; $i\leftarrow i+1$\;
  }
  \KwRet{success indicator: terminal task reward $r$}
  \end{algorithm}

\Cref{alg:roboaware-deploy} separates offline training from deployment. During training, we collect \pschema{} search roots, run MCTS from those roots, and fit $Q_\theta$ on the resulting transitions. During deployment, the trained coordinator selects a family at each eligible stage boundary and updates its input from execution feedback.
Only $Q_\theta$ is optimized. The coding agent $\codepolicy$ and API library are fixed before data collection and remain frozen during training and deployment.

\paragraph{Trajectory collection and neighboring-state sampling.}
For each training task, $\codepolicy$ constructs a task plan $\mathcal{G}$ comprising \pschema{} skill schemas, with $g_i$ identifying the active skill and $z_i$ its current stage.
A fixed sampling rule with support on all admissible families selects the family before each code block is generated.
During the resulting simulator rollout $\tau$, we record observations, execution history, schema states, and saved simulator states, together with the start time $t_i$ of each visited stage.
For each $t_i$, let $\mathcal{W}_i(\Delta)$ contain the nonterminal saved times within $\Delta$ steps of $t_i$ at which the same stage-entry conditions hold and control is available to the coding agent.
It includes $t_i$ and excludes intermediate states inside an atomic API invocation.
We draw $S$ times independently and uniformly from this set, with replacement, and form a separate root snapshot for each draw.
Each root uses its own matching context $x_t=(\ell,o_t,h_t,z_t,g_t)$. The snapshot, denoted by $\xi_t=(x_t,m_t,p_t,r)$, also stores the simulator and execution state needed to restore that root exactly. Here, $r$ denotes access to the task-success evaluator rather than a future reward available to the coordinator.
Thus, neighboring roots may differ physically, but all counterfactual branches of one root share the same physical state and observable-history prefix.

\paragraph{Search and learning contracts.}
Each root receives a budget of $K$ completed suffix simulations, including the initial family comparisons.
UCB descent follows stored tree paths using \cref{eq:ucb}, stopping before terminal nodes.
At a new node, counts start at zero and $\widehat{Q}_i(c)$ is initialized from $Q_\theta(x_i,c)$, whose parameters remain fixed during data collection.
Descent stops at an unexpanded nonterminal node when the remaining budget covers every admissible family.
Otherwise, it stops at an expanded nonterminal node and selects one family for reevaluation.
Every evaluation restores the selected snapshot, samples a fresh family-constrained code block, executes it to the next semantic boundary, and saves the resulting transition and successor snapshot.
From each nonterminal successor, the fixed coding agent and API library continue physics execution under the same family-sampling continuation rule to task termination, unrecoverable failure, the episode horizon, or an empty admissible-family set.
Reaching any of these stopping conditions sets $d=1$ and ends the suffix. States visited only during rollout are not added to the search tree.
The suffix return is backed up along the selected path using the recorded prefix rewards, with visit counts incremented and $\widehat Q$ updated by empirical averaging.
All expanded edges enter $\mathcal{D}_{\mathrm{SCB}}$, including failed and nonpreferred branches. Scores $r_{\mathrm p}$ and $r_{\mathrm m}$ are emitted at skill completion and are zero at other transitions, while $r$ is emitted only at task termination.
Q-learning then uses \cref{eq:eal-loss,eq:smooth-l1}, with no bootstrap at terminal states and a target copy refreshed every $J$ updates.
The sampling parameters $(S,\Delta)$, family-sampling rule, optimization budget $M$, learning rate $\eta$, target-update interval $J$, and related budgets are listed in~\Cref{tab:hyperparameters}.

\paragraph{Deployment contract.}
Only the trained $Q_\theta$, frozen coding agent, and API library are retained.
At every decision point, the coordinator builds $x_i$ from observable information, selects an admissible family, and conditions code generation on that choice.
Each atomic API call runs until its postcondition is satisfied, a failure is reported, or its step budget is exhausted. Once the call returns, the coding agent uses the feedback to update the history and remaining plan and to determine whether to advance or retry a stage.
Task termination, an unrecoverable failure, or the episode horizon ends execution. An empty admissible-family set is treated as failure.
Neither state restoration, tree search, oracle shaping rewards, nor parameter updates are used at deployment. Only the terminal task reward $r$ is reported.

\paragraph{Training data and task pool.}
We construct $\mathcal{D}_{\mathrm{SCB}}$ using the procedure in~\Cref{alg:roboaware-deploy} and the model in~\Cref{app:q-model-details}, with the sampling and optimization settings given below.
Data collection with privileged simulator access and subsequent $Q_\theta$ updates use only scene seed~$0$ on the coordinator training pool ($80$ LIBERO-Pro, $4$ RoboSuite, and $16$ RoboTwin~2.0 tasks;~\Cref{app:benchmark-details}).
Each training task contributes (i)~one or more physics trajectories used to mark \pschema{} stage-entry times, (ii)~$S$ neighboring roots per stage entry drawn from $\mathcal{W}_i(\Delta)$ as specified above, and (iii)~up to $K$ completed suffix simulations per root, with every expanded family edge written into $\mathcal{D}_{\mathrm{SCB}}$.

\paragraph{Root and family sampling.}
During trajectory collection, we sample uniformly from the families admissible at the current $(z_i,g_i)$, so every family in $\mathcal{C}_i$ has nonzero probability.
Neighboring-state sampling draws $S{=}4$ roots independently and uniformly with replacement from $\mathcal{W}_i(\Delta)$ using the default window $\Delta{=}10$ saved control steps.
Within each root budget of $K{=}20$ completed suffix simulations (\Cref{tab:hyperparameters}), the first expansion of a new node evaluates every $c\in\mathcal{C}_i$ once via~\scb{}; later visits reevaluate a single UCB-selected family (\cref{eq:ucb}).
Rollout continuation after expansion reuses the same family-sampling rule and does not add rollout-only states as tree nodes.
All evaluated edges, including failed and nonpreferred branches, remain in $\mathcal{D}_{\mathrm{SCB}}$.

\paragraph{Q initialization before data collection.}
Before data collection, we load the released Qwen3.5-0.8B backbone weights, randomly initialize the \texttt{<VALUE>} embedding and MLP head, and set $\bar\theta\leftarrow\theta$ (\Cref{alg:roboaware-deploy}).
We then hold $\theta$ fixed throughout collection. Each new tree node starts with zero visits and initializes $\widehat{Q}_i(c)$ from $Q_\theta(x_i,c)$, which provides the initial value estimate for UCB selection.
No return regression or supervised fine-tuning is performed before collection, and the model receives no privileged oracle features. Bellman updates begin only after $\mathcal{D}_{\mathrm{SCB}}$ is complete, following the single collection-and-fitting cycle used in the reported experiments.

\paragraph{Optimization protocol.}
After collection, we optimize $\theta$ on the fixed offline set with the Smooth-$L_1$ Bellman loss in~\cref{eq:eal-loss,eq:smooth-l1}, minibatches drawn from $\mathcal{D}_{\mathrm{SCB}}$, a stop-gradient target network $Q_{\bar\theta}$, and no bootstrap at terminal states.
\Cref{tab:q-opt-settings} lists the fine-tuning method, thinking setting, optimizer, batch size, learning-rate schedule (cosine with warmup), update budget $M$, target refresh interval $J$, numeric precision, and checkpoint rule.
We select a checkpoint on the validation split described below, following the checkpoint rule in~\Cref{tab:q-opt-settings}, and freeze it for all evaluations on seeds~$1$--$10$.

\begin{table}[H]
\centering
\caption{Coordinator Q-fitting settings. Symbols match~\Cref{alg:roboaware-deploy}.}
\label{tab:q-opt-settings}
\small
\begin{tabular}{@{}p{0.48\linewidth}p{0.42\linewidth}@{}}
\toprule
Setting & Value \\
\midrule
Fine-tuning method & full-FT (backbone + MLP value head) \\
Thinking & Disabled \\
Optimizer & AdamW \\
Minibatch size & $16$ \\
Learning rate $\eta$ / schedule & $10^{-5}$, cosine, $5\%$ warmup \\
Gradient updates $M$ & $10{,}000$ \\
Target refresh every $J$ updates & $500$ (hard copy $\bar\theta\leftarrow\theta$) \\
Weight decay / grad clip & $10^{-4}$ / $1.0$ \\
Numeric precision & bf16 \\
Checkpoint selection & lowest validation Bellman loss \\
\bottomrule
\end{tabular}
\end{table}

\paragraph{Grouped training and validation split.}
The $\modfamily$ and $\eetfamily$ branches from a given root share both a physical state and an observable-history prefix. Roots sampled from the same neighborhood $\mathcal{W}_i(\Delta)$ are also closely related because they come from nearby states in one trajectory.
A random split of individual transitions could therefore place branches from the same root, or transitions from neighboring roots, in both training and validation.
To keep these related samples together, we group all transitions from each root and all roots drawn from the same stage-entry neighborhood $\mathcal{W}_i(\Delta)$.
Groups are assigned entirely to training or validation using a $90/10$ split, optionally stratified by benchmark and by whether $\mathcal{C}_i$ is singleton or binary.
No group contributes transitions to both splits; evaluation seeds~$1$--$10$ remain unused for fitting or model selection.

\subsection{Responsibility Preference}\label{app:responsibility-preference}
The normalization in \cref{eq:responsibility-preference} defines a relative preference over the nonempty admissible-family set $\mathcal C_i$.
When $\mathcal C_i=\{\eetfamily,\modfamily\}$, the value gap compares the same ordered pair at every decision point:
\begin{equation}
\label{eq:responsibility-gap}
\Delta_\theta(x_i)=Q_\theta(x_i,\eetfamily)-Q_\theta(x_i,\modfamily).
\end{equation}
Thus, $\Delta_\theta(x_i)$ depends on the context, while the family identities and their order are fixed by definition.
The normalized preferences can then be expressed as sigmoids of this gap. Dividing the E2E numerator and denominator by $\exp(Q_\theta(x_i,\eetfamily))$ gives
\begin{equation}
\label{eq:responsibility-sigmoid}
\begin{aligned}
\rho_\theta(\eetfamily\mid x_i)
&=\frac{\exp(Q_\theta(x_i,\eetfamily))}
{\exp(Q_\theta(x_i,\eetfamily))+\exp(Q_\theta(x_i,\modfamily))}\\
&=\frac{1}{1+\exp(Q_\theta(x_i,\modfamily)-Q_\theta(x_i,\eetfamily))}\\
&=\sigma(\Delta_\theta(x_i)),\\
\rho_\theta(\modfamily\mid x_i)
&=1-\rho_\theta(\eetfamily\mid x_i)
=\sigma(-\Delta_\theta(x_i)).
\end{aligned}
\end{equation}
The common normalization cancels in the ratio, so taking its logarithm yields
\begin{equation}
\label{eq:responsibility-log-ratio}
\begin{aligned}
\log\frac{\rho_\theta(\eetfamily\mid x_i)}{\rho_\theta(\modfamily\mid x_i)}
&=\log\frac{\exp(Q_\theta(x_i,\eetfamily))}{\exp(Q_\theta(x_i,\modfamily))}\\
&=Q_\theta(x_i,\eetfamily)-Q_\theta(x_i,\modfamily)
=\Delta_\theta(x_i).
\end{aligned}
\end{equation}
Positive gaps favor E2E, negative gaps favor modular composition, and a zero gap gives equal preference to both families.
The larger preference equals $\sigma(|\Delta_\theta(x_i)|)$ and increases monotonically with $|\Delta_\theta(x_i)|$, which therefore measures preference strength.

More generally, exponentiation is strictly increasing and the denominator in \cref{eq:responsibility-preference} is positive and shared by all admissible families.
Consequently, the transformation preserves family ordering, including ties, and yields the same maximizing families:
\begin{equation}
\label{eq:responsibility-ordering}
\arg\max_{c\in\mathcal C_i}\rho_\theta(c\mid x_i)
=\arg\max_{c\in\mathcal C_i}Q_\theta(x_i,c).
\end{equation}
When only one family is admissible, its normalized preference is one; the binary sigmoid identities apply only when both families are admissible.
An empty admissible-family set terminates execution with failure, as specified in the deployment phase of \Cref{alg:roboaware-deploy}.
The deployment algorithm therefore selects via $\rho_\theta$ (equivalently via $Q_\theta$), then generates and executes the corresponding block before reconsidering responsibility at the next semantic boundary.

\subsection{Q-Value Model}\label{app:q-model-details}
We instantiate $Q_\theta(x_i,c)$ with Qwen3.5-0.8B as the multimodal backbone and a scalar MLP value head.
Only this coordinator value model is optimized under~\cref{eq:eal-loss}; the coding agent $\codepolicy$, the typed API library, and all low-level backend checkpoints remain frozen throughout \eal{} and deployment.

\paragraph{Observation encoding.}
At decision point $i$, the visual stream uses the same agent-view camera as the modular perception API.
We resize the agent-view RGB frame $I_i^{\mathrm{rgb}}$ and a visualization of the metric depth map $I_i^{\mathrm{d}}$ to a shared resolution $R{\times}R$ (\Cref{tab:q-model-settings}) and encode both through the backbone's native image pathway.
We serialize proprioception $q_i$, including joint positions, end-effector pose, and gripper opening, as structured text.
Thus, all components of $o_i$ enter $Q_\theta$ through its image and text pathways, without a separate numeric depth input, depth adapter, or proprioceptive MLP encoder.

\paragraph{History length and truncation.}
The execution history $h_i$ retains the most recent $H$ stage transitions in full, each recording the chosen family, typed API returns or failure codes, stage advances or retries, and remaining budgets.
Older transitions are dropped and replaced by a fixed-length summary of completed skills/stages, persistent object bindings, cumulative failure counts, and the residual step/wall-clock budget.
The summarizer uses only information already present in the observable history: training-time oracle fields never enter $h_i$.

\paragraph{Token order.}
For a queried family $c$, inputs are packed into one multimodal sequence:
\begin{equation*}
\ell \;\rightarrow\;
(I_i^{\mathrm{rgb}},I_i^{\mathrm{d}}) \;\rightarrow\;
q_i \;\rightarrow\;
h_i \;\rightarrow\;
g_i \;\rightarrow\;
z_i \;\rightarrow\;
c \;\rightarrow\;
\texttt{<VALUE>},
\end{equation*}
i.e., instruction, images, proprioception, execution history, current skill, current \pschema{} stage, family identity, then the special \texttt{<VALUE>} token.
For example, consider RoboTwin~\texttt{place\_bread\_basket} at the start of the \textit{perform} stage for picking up the bread. The instruction $\ell$ is ``Place the bread into the basket,'' and the image inputs show the current $R{\times}R$ RGB frame and depth colormap of the bread, basket, and two arms. The proprioceptive input $q_i$ lists joint positions, end-effector poses, and gripper openings.
The history $h_i$ records the preceding stage transitions. These may include $\modfamily$ returning the bread pose at \textit{perceive}, producing an end-effector target at \textit{propose}, and reaching the pre-grasp pose at \textit{pre-manipulate}. It also contains the remaining step and wall-clock budgets and a fixed-length summary of earlier completed skills.
The active schema $g_i$ identifies the bread-picking skill and its unmet postconditions, while $z_i=\textit{perform}$. Both families are admissible at this point, so the sequence ends with the queried $c\in\{\modfamily,\eetfamily\}$ followed by \texttt{<VALUE>}.
The backbone produces the final-layer hidden state at \texttt{<VALUE>}; an $L_{\mathrm{mlp}}$-layer MLP with hidden width $d_{\mathrm{mlp}}$ maps that state to the scalar $Q_\theta(x_i,c)$.
\Cref{tab:q-model-settings} lists the quantitative settings used in our experiments.

\begin{table}[H]
\centering
\caption{Q-value model settings.}\label{tab:q-model-settings}
\small
\begin{tabular}{@{}p{0.48\linewidth}p{0.42\linewidth}@{}}
\toprule
Setting & Value \\
\midrule
Backbone & Qwen3.5-0.8B \\
Image viewpoint & agent-view RGB + depth visualization \\
Image resolution $R{\times}R$ & $224{\times}224$ \\
Proprioception encoding & structured text \\
Depth encoding & colormap image \\
History window $H$ (full transitions) & $4$ \\
MLP depth $L_{\mathrm{mlp}}$ & $2$ \\
MLP width $d_{\mathrm{mlp}}$ & $512$ \\
Frozen parameters & $\codepolicy$, API library, backend checkpoints \\
\bottomrule
\end{tabular}
\end{table}

\subsection{Reward Function}\label{app:reward-details}
This subsection specifies the training-only oracle rewards in \cref{eq:transition-reward}.
The coordinator never observes \(m_i\), \(p_i\), \(r_{\mathrm p}\), \(r_{\mathrm m}\), or \(r\) at deployment.
As in \cref{app:deployment-algo}, \(r_{\mathrm{p},i}^c\) and \(r_{\mathrm{m},i}^c\) are emitted when a \pschema{} skill completes under family \(c\) and are zero on non-completion transitions; the binary task reward \(r\) is emitted only when the episode terminates (\(d_i^c=1\)).

\paragraph{Perception score \(r_{\mathrm p}\).}
Let \(m_i\) be the oracle binary mask of the language-referenced target object in the agent-view image at the skill-entry snapshot \(\xi_i\).
When the executed block exposes an explicit mask \(\hat m_i^c\) (modular \texttt{Perceive\_object\_pose}), mask agreement is the intersection-over-union
\begin{equation}
\label{eq:mask-iou}
r_{\mathrm{p},i}^c
=
\mathrm{IoU}(\hat m_i^c,m_i)
=
\frac{|\hat m_i^c\cap m_i|}{|\hat m_i^c\cup m_i|}\in[0,1],
\end{equation}
with the convention \(\mathrm{IoU}=0\) if both masks are empty.
End-to-end blocks do not return a mask.
For those branches we score target-object identity rather than pixel overlap: after execution, the simulator identifies the object instance that was contacted or whose pose changed under the skill; \(r_{\mathrm{p},i}^c=1\) if that instance matches the object labeled by \(m_i\), and \(0\) otherwise.
Thus both families receive a perception score in \([0,1]\), but only modular perception is judged by mask IoU; E2E perception is judged by whether the call acted on the oracle-referenced object, consistent with the E2E postcondition coverage in \Cref{tab:stage-contracts}.

\paragraph{Manipulation score \(r_{\mathrm m}\).}
The manipulation score measures whether the object's simulator ground-truth pose after skill execution meets the skill's geometric goal.
Write \(p_{i+1}^c\) for the privileged 6-DoF object pose read from \(\xi_{i+1}^c\), and \(p_i^\star\) for the skill goal pose supplied by the same oracle pose channel (placement target, handover target, or other stage goal recorded with the task).
Let \(t(\cdot)\) and \(R(\cdot)\) denote translation and rotation.
We measure translation error in meters using the Euclidean distance \(\|t(p_{i+1}^c)-t(p_i^\star)\|_2\), and rotation error in radians using the geodesic angle on \(\mathrm{SO}(3)\),
\(\angle\!\bigl(R(p_{i+1}^c),R(p_i^\star)\bigr)=\arccos\bigl((\mathrm{tr}(R(p_{i+1}^c)^\top R(p_i^\star))-1)/2\bigr)\).
For skills defined by a contact or grasp postcondition, such as a successful lift, \(r_{\mathrm{m},i}^c\) is the corresponding binary oracle predicate. For skills with a 6-DoF placement goal, we instead apply the pose thresholds
\begin{equation}
\label{eq:pose-success}
r_{\mathrm{m},i}^c
=
\mathbb{I}\!\left[
\|t(p_{i+1}^c)-t(p_i^\star)\|_2\le\epsilon_p
\;\wedge\;
\angle\!\bigl(R(p_{i+1}^c),R(p_i^\star)\bigr)\le\epsilon_R
\right],
\end{equation}
where \(\epsilon_p\) is the translational threshold in meters and \(\epsilon_R\) is the rotational threshold in radians (\Cref{tab:hyperparameters}).
The main-text phrase ``object-pose error against \(p_i\)'' refers to this privileged pose channel used to form \((p_{i+1}^c,p_i^\star)\), not to scoring the modular pose estimate against the entry state.

\paragraph{Terminal reward and stopping.}
The evaluator returns \(r=1\) iff the task success predicate holds at termination, and \(r=0\) otherwise.
An episode sets \(d_i^c=1\) and stops on task success, unrecoverable failure, empty admissible-family set, or the episode horizon; only the success case can yield \(r=1\).
The shaped transition reward remains \(\tilde r_i^c=w_{\mathrm p}r_{\mathrm{p},i}^c+w_{\mathrm m}r_{\mathrm{m},i}^c+w_{\mathrm t}d_i^cr\) with \(w_{\mathrm p}+w_{\mathrm m}+w_{\mathrm t}=1\).

\begingroup
\raggedbottom
\renewenvironment{table*}[1][]{\begin{table}[!htbp]}{\end{table}}

\section{API Library Instantiation}
\label{app:embodied-apis}
\suppressfloats[t]

\subsection{Typed API Contracts}
\label{app:api-contracts}

Each embodied API in \mainCref{tab:embodied-apis} owns one or more \pschema{}
stages and declares typed inputs, outputs, coordinate frames, postconditions,
and failure states (\Cref{tab:embodied-apis-io}).  The family-specific rules in
\Cref{tab:family-api-library} determine which APIs a generated code block
may call.  The coding agent specifies the semantic target and route, whereas
embodiment adapters resolve frames, execute the request, and report measured
completion.  This separates an incorrect target from a failure to reach a
correct target and prevents an internal fallback from silently changing the
meaning of a call.

Within $\mathcal{A}_{\mathrm{E2E}}$, \texttt{route\_id} selects a retained VLA,
video-action, or world-action checkpoint inside the code block; it is not a
separate action scored by $Q_\theta$.  The coordinator observes only typed API
returns, not privileged simulator state.  We freeze the coding agent, API
contracts, backend bindings, and checkpoints before coordinator training.

\begin{table*}[t]
\centering
\caption{
Typed input/output contracts for the high-level embodied APIs in \mainCref{tab:embodied-apis}.
The E2E route uses \texttt{route\_id} to select one frozen backend.
}
\label{tab:embodied-apis-io}
\resizebox{\textwidth}{!}{%
\begin{tabular}{llll}
\toprule
API & Stage & Input & Output \\
\midrule
\texttt{Perceive\_object\_pose} & \textit{perceive} & task instruction + agent-view RGB-D + calibration & target pose + mask/geometry provenance \\
\texttt{Propose\_eef\_pose} & \textit{propose} & RGB-D geometry + target mask + robot state + task context & ranked end-effector target + screening provenance \\
\texttt{Plan\_to\_eef\_pose} & \textit{pre}/\textit{post-manipulate} & scene observation, robot state, target pose, optional attachment & continuous, smooth, collision-aware qpos sequence \\
\texttt{Perform\_gripper} & \textit{perform} & gripper open/close command & updated gripper state \\
\performendtoend & \makecell[l]{\textit{perceive}, \textit{pre-manipulate},\\ \textit{perform}, \textit{post-manipulate}} & route ID + task instruction + RGB-D + proprioception & closed-loop control until postcondition, failure, or budget \\
\bottomrule
\end{tabular}%
}
\end{table*}

\begin{table*}[t]
\centering
\caption{
Policy-family bindings for generated code blocks.
Mod composes stage-admissible APIs; E2E invokes one frozen backend through \texttt{route\_id}.
}\label{tab:family-api-library}
\resizebox{\textwidth}{!}{%
\begin{tabular}{llll}
\toprule
Policy family & Code-block space & Permitted API pattern & Backend choice \\
\midrule
\modfamily & $\mathcal{A}_{\mathrm{Mod}}$ & stage-admissible modular API composition & fixed selected modular backends \\
\eetfamily & $\mathcal{A}_{\mathrm{E2E}}$ & \performendtoend{}$(\texttt{route\_id},\ldots)$ & retained frozen backend selected inside $a_i$ \\
\bottomrule
\end{tabular}%
}
\end{table*}

\subsection{Object Perception}
\label{app:perceive}
\texttt{Perceive\_object\_pose} grounds a language reference in an agent-view
RGB-D observation and combines the resulting mask with camera calibration and
depth to recover a robot-frame target.  Valid masked depth pixels are
back-projected with the camera intrinsics, transformed through the camera
extrinsics, filtered for invalid or isolated points, and summarized by a
robust centroid and oriented extent.  The API returns the pose together
with the mask, supporting points, confidence, and source observation.
If the mask is empty, the depth is inconsistent, or the reference is ambiguous,
it reports a failure instead of returning a pose.

We use InstructSAM~\citep{InstructSAM} as the grounding backend because its
instruction-conditioned interface directly accepts the detailed free-form
prompt produced by the coding agent from the current task context.  The agent
can therefore preserve task-specific attributes and spatial relations in a
single perception call.  The resulting mask is lifted through the shared
geometric pipeline, so downstream proposal and motion APIs retain the same
typed outputs and failure checks.

\subsection{Pose Proposal}
\label{app:propose}
\texttt{Propose\_eef\_pose} consumes the frozen RGB-D observation, target mask,
and derived geometry and returns a grasp target together with its score, rank,
predicted width, coordinate transform, screening result, and provenance.  We
compare proposal methods using the same cuRoboV2 downstream planner and retain
each baseline's original candidate ranking. With the planner held fixed,
Success@$k$ in \Cref{tab:grasp_proposal_comparison} reflects differences in
candidate coverage and ordering.

This API owns grasp proposal, not placement-pose generation.  For placement,
the task supplies a target object pose; the motion API resolves it to a TCP
target through the held-object transform and ranks routes by feasibility,
continuity, and the declared approach or contact constraints.

Our NeuGraspNet+ backend addresses two failure modes of native NeuGraspNet
\citep{NeuGraspNet}.  When native geometric sampling produces too few
candidates, a \emph{RobustSampler} derives deterministic top-down, side, and
oblique seeds from the target point cloud and its principal directions.  The
released NeuGraspNet checkpoint still predicts their quality and gripper
width, so the fallback expands coverage without another grasp network or
task-specific proposals.

NeuGraspNet+ then uses cuRoboV2 to identify up to five candidates with feasible
approach, grasp, and lift trajectories, and selects the top-ranked grasp from
this feasible set.  Its confidence gate preserves the network ranking when scores
are informative, uses geometry to break near-ties, and gives geometry greater
weight when all feasible scores are low.  This avoids globally replacing a
useful learned ranking with a fixed geometric prior.

NeuGraspNet+ reaches $18/18$ Success@1, Success@3, and Success@5 on the
18-task Panda screen (\Cref{tab:grasp_proposal_comparison}) and is frozen as
the grasp backend.  These selection rules prioritize first-choice grasp success over speed and
do not involve retraining NeuGraspNet. Because they were developed using the
benchmark, their performance on new object distributions requires further
validation.

\begin{table*}[t]
  \centering
  \caption{Grasp proposal results on the 18-task Panda benchmark. All baseline
  methods use their native proposal rankings; NeuGraspNet uses our final
  selection pipeline. All methods share the same downstream cuRoboV2 motion
  planner. Success@$k$ indicates whether a successful grasp is found within
  the first $k$ evaluated proposals. Proposal and screening times are averaged
  over all 18 tasks; Time@1 is averaged over successful Top-1 trials.}
  \label{tab:grasp_proposal_comparison}
  \small
  \setlength{\tabcolsep}{5pt}
  \resizebox{\linewidth}{!}{%
  \begin{tabular}{lcccccc}
    \toprule
    Method
      & Success@1 $\uparrow$
      & Success@3 $\uparrow$
      & Success@5 $\uparrow$
      & Proposal time (s) $\downarrow$
      & Screening time (s) $\downarrow$
      & Time@1 (s) $\downarrow$ \\
    \midrule
    AnyGrasp~\citep{AnyGrasp}      & \underline{13/18} & \underline{16/18} & \underline{17/18} & 7.98  & 3.46 & 36.39 \\
    GraspGen~\citep{GraspGen}      & 11/18             & 12/18             & 12/18             & \textbf{4.54} & 3.16 & \underline{24.84} \\
    NeuGraspNet~\citep{NeuGraspNet}   & 12/18             & 13/18             & 13/18             & 12.90 & 1.56 & 39.05 \\
    EconomicGrasp~\citep{EconomicGrasp} & 10/18             & 12/18             & 12/18             & \underline{5.03} & \textbf{1.11} & \textbf{20.68} \\
    RNGNet~\citep{RNGNet}        & 7/18              & 8/18              & 8/18              & 18.89 & \underline{1.34} & 46.33 \\
    \midrule
    NeuGraspNet+ (ours)     & \textbf{18/18}    & \textbf{18/18}    & \textbf{18/18}    & 14.36 & 2.23 & 49.04 \\
    \bottomrule
  \end{tabular}%
  }
\end{table*}

\subsection{Motion Planning}
\label{app:plan-motion}
\texttt{Plan\_to\_eef\_pose} converts a caller-specified end-effector or
held-object target into the robot's command frame, executes a trajectory, and
verifies arrival.  The fixed-grasp protocol in
\Cref{tab:motion_planning_comparison} supplies every planner with the same
target and therefore isolates planning, whereas planner-native results measure
the complete native route.  We retain cuRoboV2~\citep{cuRoboV2} as the core
global planner and trajectory checker; it is also shared by the proposal
comparison in \Cref{tab:grasp_proposal_comparison}.

For each checked call, the API freezes the current RGB-D frame, robot state,
camera calibration, and target mask.  It filters robot-image residuals and
excludes the grasp target where contact is intended.  Closing a successful
grasp creates a held-object state containing the measured hand--object
transform and attachment geometry; later paths check both robot and attachment
against the environment.  With \texttt{target\_frame=``tcp''}, the public TCP
goal is converted to the native hand frame.  With
\texttt{target\_frame=``held\_object''}, the saved transform determines the hand
goal that realizes a desired object pose.

\paragraph{Explicit motion modes.}
\texttt{goto\_pose} provides three motion modes with different planning costs
and collision checks.  The default
\texttt{auto} mode uses cuRoboV2 first, collision-checked multi-seed IK second,
and the existing bounded recovery route last; it is appropriate for grasping,
long transport, and cluttered motion.  \texttt{ik\_checked} skips global
planning but captures the scene and collision-checks dense IK branches,
including the attachment, for short contact-free adjustments.
\texttt{ik\_unchecked} skips scene capture and scene-collision checking for
deliberate contact such as staged insertion, drawer closing, or door pushing,
while retaining multi-seed IK, strict joint-continuity limits, dense execution,
tracking, and arrival checks.  The ``unchecked'' mode therefore omits scene-collision checks while retaining
these control and execution checks.

The mode is selected before execution so that intended contact does not
consume the budget in repeated collision-avoidance attempts. This also ensures
that a call requested with collision checking cannot silently switch to
unchecked execution.  For example, a microwave
program can use \texttt{auto} for grasp and transport,
\texttt{ik\_checked} for final free-space alignment, and
\texttt{ik\_unchecked} for insertion or a door-pushing arc.

\paragraph{Approach and execution acceptance.}
\texttt{z\_approach} adds a waypoint along the declared tool frame.  For
held-object placement, \texttt{contact\_target} marks only a bounded terminal
segment as expected contact, leaving the preceding transport checked.  Before
returning, every route monitors joint tracking and verifies joint and
end-effector arrival.  The typed result records the resolved frame, motion
mode, trajectory source, performed checks, timing, residuals, and a specific
failure category, allowing history $h_i$ to distinguish planning, execution,
and semantic-target failures.

\begin{table*}[h]
  \centering
  \caption{Motion-planning results on the 18-task Panda benchmark. The
  planner-native fresh protocol is the main end-to-end comparison: each
  planner independently resolves endpoints, screens proposals, plans, and
  executes. Pipeline time includes candidate screening, endpoint resolution,
  and planning; Plan time measures the final selected trajectory only. The
  fixed-grasp protocol uses a shared grasp target. The
  conditional diagnostic uses cuRobo-success anchors and should not be ranked
  directly against the planner-native results. Avoid Everything was evaluated
  only in the conditional pose-goal diagnostic.}
  \label{tab:motion_planning_comparison}
  \small
  \setlength{\tabcolsep}{2.5pt}
  \resizebox{\textwidth}{!}{%
  \begin{tabular}{lccccccc}
    \toprule
    & \multicolumn{3}{c}{Planner-native fresh (main)}
    & \multicolumn{2}{c}{Fixed grasp}
    & \multicolumn{2}{c}{Conditional diagnostic} \\
    \cmidrule(lr){2-4}\cmidrule(lr){5-6}\cmidrule(lr){7-8}
    Planner
      & Plannable $\uparrow$
      & Task success $\uparrow$
      & Pipeline / Plan (s) $\downarrow$
      & Task success $\uparrow$
      & Plan (s) $\downarrow$
      & Grasp success $\uparrow$
      & Plan (s) $\downarrow$ \\
    \midrule
    cuRoboV2~\citep{cuRoboV2}          & \textbf{18/18} & \textbf{18/18} & \textbf{3.82 / 0.254} & \textbf{15/18} & \textbf{0.192} & \textbf{18/18} & \textbf{0.935} \\
    DRP/IMPACT~\citep{DRP}        & 16/18          & 14/18          & 6.23 / 0.418          & 11/18          & \underline{0.576} & 16/18 & \underline{1.526} \\
    VAMP~\citep{VAMP}              & 12/18          & 11/18          & 73.30 / 9.320         & 6/18           & 0.931 & 14/18 & 1.598 \\
    Avoid Everything~\citep{AvoidEverything}  & --             & --             & --                     & --             & --    & 2/18  & 3.677 \\
    \bottomrule
  \end{tabular}%
  }
\end{table*}

\subsection{Gripper Control}
\label{app:perform-gripper}

\texttt{Perform\_gripper} implements binary open or close actions.  A command
returns only after the observed opening or joint signal remains stable for a
fixed dwell window; failure to settle within the budget is reported explicitly.
A successful close also initializes the held-object state described above from
the measured hand pose.

The coding agent uses the same gripper interface across Franka, RoboSuite, and
RoboTwin~2.0, with platform-specific adapters mapping commands to native gripper
coordinates.  An arm
selector is omitted for single-arm platforms and explicit on bimanual systems,
preserving the selected arm through motion and release.  Because this operation
is deterministic and embodiment-local, we do not screen alternative gripper
controllers, and $Q_\theta$ does not score the adapter.

\subsection{End-to-End Policy Execution}
\label{app:perform-vla}

\performendtoend{} maps an explicit \texttt{route\_id} to a frozen VLA,
video-action, or world-action checkpoint. It supports three mutually exclusive
instruction modes. \emph{Task mode} forwards the environment task instruction
when neither an object nor a prompt is provided. \emph{Grasp--lift mode} takes
an object name, generates a local grasp-and-lift instruction, and updates the
held-object state after perception confirms an increase in the object's height.
\emph{Bounded-prompt mode} forwards an explicit prompt verbatim and requires a
step budget because arbitrary local instructions have no common success
detector. Invalid argument combinations and missing required task language
are rejected.

Execution ends on task completion, environment termination, a supported
model stop signal, or budget exhaustion; grasp--lift mode also ends when its
local completion predicate is met. Each call records the instruction and its
source, route, executed steps, termination reason, and measured postcondition.
Full-task success and grasp--lift success use the task and local predicates,
respectively.

We screen released routes on 14 LIBERO-Pro tasks, four RoboSuite tasks, and
ten RoboTwin~2.0 tasks. The screen specifies one initial state per task, but
RoboTwin's full-task evaluator rejects states that fail expert validation;
realized initial states may therefore differ across methods. Full-task and
grasp--lift success are reported in~\Cref{tab:vla_wam_environment_results},
and execution statistics in
\Cref{tab:vla_wam_efficiency_libero,tab:vla_wam_efficiency_robosuite,tab:vla_wam_efficiency_robotwin}.
Successful-trial completion times are reported separately from all-trial
runtime statistics. Retained routes receive stable \texttt{route\_id}s and
are frozen before coordinator training.

FastWAM and FasterWAM use their LIBERO checkpoints on LIBERO-Pro and RoboSuite,
and their RoboTwin checkpoints on RoboTwin. Harness-VLA uses the same
LIBERO-trained checkpoint on all three suites. The RynnVLA-002 results reflect
a loader that reinitializes part of the VQ encoder, rather than an unmodified
checkpoint deployment. JointWAM is excluded because no runnable checkpoint
was available in the audited release. These results characterize the
specified checkpoint--suite configurations.

\begin{table}[H]
  \centering
  \caption{
  Success counts across three evaluation suites.
  \textbf{Full} measures successful completion of the entire task, whereas
  \textbf{G--L} measures whether the policy successfully grasps and lifts the
  target object, providing a partial-task measure of manipulation capability.
  Each entry reports the number of successful evaluations over the total
  number evaluated; the best result in each column is shown in bold.
  $^{\star}$The $\pi$0.5-RLinf-SFT policy uses the same LIBERO-trained
  checkpoint across all three suites, without suite-specific adaptation.
  $^{\ddagger}$For RynnVLA-002, the current public loader reinitializes part of
  the VQ encoder when loading the released checkpoint.
}
  \label{tab:vla_wam_environment_results}
  \small
  \setlength{\tabcolsep}{5.0pt}
  \resizebox{\linewidth}{!}{%
  \begin{tabular}{lcccccc}
    \toprule
    & \multicolumn{2}{c}{LIBERO-Pro}
    & \multicolumn{2}{c}{RoboSuite}
    & \multicolumn{2}{c}{RoboTwin 2.0} \\
    \cmidrule(lr){2-3}
    \cmidrule(lr){4-5}
    \cmidrule(lr){6-7}
    Method & Full & G--L & Full & G--L & Full & G--L \\
    \midrule

    \multicolumn{7}{l}{\textit{Vision-Language-Action Policies}} \\
    \addlinespace[1pt]

    OpenVLA-OFT~\citep{OpenVLAOFT}
      & 4/14 & 5/14
      & 0/4 & 0/4
      & 0/10 & 1/10 \\

    PRTS~\citep{PRTS}
      & 6/14 & 7/14
      & \textbf{1/4} & 1/4
      & 0/10 & 4/10 \\

    RDT2~\citep{RDT2}
      & 0/14 & 0/14
      & 0/4 & 0/4
      & 0/10 & 0/10 \\

    SpatialVLA~\citep{SpatialVLA}
      & 0/14 & 0/14
      & 0/4 & 0/4
      & 0/10 & 1/10 \\

    StarVLA~\citep{StarVLA}
      & 4/14 & 3/14
      & 0/4 & 0/4
      & 0/10 & 2/10 \\

    RynnVLA-002$^{\ddagger}$~\citep{RynnVLA002}
      & 1/14 & 3/14
      & 0/4 & 0/4
      & 0/10 & 1/10 \\

    EVO1~\citep{EVO1}
      & 4/14 & 3/14
      & 0/4 & 0/4
      & 0/10 & 3/10 \\

    ABot-M0~\citep{ABotM0}
      & 4/14 & 3/14
      & \textbf{1/4} & 0/4
      & 7/10 & 9/10 \\

    InternVLA-A1.5~\citep{InternVLAA15}
      & 5/14 & 7/14
      & 0/4 & 0/4
      & 9/10 & 9/10 \\

    SmolVLA~\citep{SmolVLA}
      & 3/14 & 4/14
      & 0/4 & 0/4
      & 0/10 & 3/10 \\

    WLA-0~\citep{WLA0}
      & 4/14 & 4/14
      & \textbf{1/4} & 1/4
      & 8/10 & \textbf{10/10} \\

    X-VLA~\citep{XVLA}
      & 2/14 & 4/14
      & 0/4 & 0/4
      & 0/10 & 0/10 \\

    MolmoAct2~\citep{MolmoAct2}
      & \textbf{10/14} & 11/14
      & \textbf{1/4} & \textbf{2/4}
      & 0/10 & 0/10 \\

    $\pi$0.5-RLinf-SFT$^{\star}$~\citep{pi05}
      & \textbf{10/14} & \textbf{13/14}
      & 0/4 & \textbf{2/4}
      & 0/10 & 2/10 \\

    \midrule

    \multicolumn{7}{l}{\textit{Video-Action Models}} \\
    \addlinespace[1pt]

    LingBot-VA~\citep{LingBotVA}
      & 0/14 & 3/14
      & 0/4 & \textbf{2/4}
      & 9/10 & 9/10 \\

    \midrule

    \multicolumn{7}{l}{\textit{World-Action Models}} \\
    \addlinespace[1pt]

    FastWAM~\citep{FastWAM}
      & 4/14 & 11/14
      & 0/4 & 0/4
      & \textbf{10/10} & \textbf{10/10} \\

    FasterWAM~\citep{FasterWAM}
      & 4/14 & 11/14
      & 0/4 & 0/4
      & 8/10 & \textbf{10/10} \\

    \bottomrule
  \end{tabular}%
  }
\end{table}
{\setlength{\intextsep}{4pt plus 1pt minus 1pt}%
 \captionsetup{skip=2pt}%
\begin{table}[H]
  \centering
  \caption{
  Execution efficiency on LIBERO-Pro under the \textbf{Full} and \textbf{G--L}
  criteria. \textbf{Full} denotes end-to-end task success, while
  \textbf{G--L} denotes successful grasp and lift of the target object.
  For each criterion, \textbf{All trials} aggregates both successful and failed
  episodes and reports the average number of executed steps, average execution
  time, and effective control frequency.
  \textbf{Successful only} reports the mean and median completion time over
  successful episodes only.
  A dash indicates that the method achieves no successful trial under that
  criterion and therefore has no successful-only timing statistic.
  Lower all-trial steps or execution time do not necessarily indicate greater
  efficiency, since failed episodes may terminate before completing the full
  rollout horizon; these quantities should therefore be interpreted jointly
  with the corresponding success counts in
  \Cref{tab:vla_wam_environment_results}.
}
  \label{tab:vla_wam_efficiency_libero}

  \footnotesize
  \setlength{\tabcolsep}{2.0pt}
  \renewcommand{\arraystretch}{1.05}

  \resizebox{\linewidth}{!}{%
  \begin{tabular}{@{}l
    S[table-format=3.1]
    S[table-format=4.1]
    S[table-format=1.3]
    S[table-format=3.1]
    S[table-format=3.1]
    S[table-format=3.1]
    S[table-format=4.1]
    S[table-format=1.3]
    S[table-format=3.1]
    S[table-format=3.1]@{}}

    \toprule
    & \multicolumn{5}{c}{Full}
    & \multicolumn{5}{c}{G--L} \\
    \cmidrule(lr){2-6}
    \cmidrule(lr){7-11}

    & \multicolumn{3}{c}{All trials}
    & \multicolumn{2}{c}{Successful only}
    & \multicolumn{3}{c}{All trials}
    & \multicolumn{2}{c}{Successful only} \\
    \cmidrule(lr){2-4}
    \cmidrule(lr){5-6}
    \cmidrule(lr){7-9}
    \cmidrule(lr){10-11}

    Method
      & {Steps}
      & {Time (s)}
      & {Hz}
      & {Mean (s)}
      & {Median (s)}
      & {Steps}
      & {Time (s)}
      & {Hz}
      & {Mean (s)}
      & {Median (s)} \\
    \midrule

    \multicolumn{11}{@{}l}{
      \textit{Vision--Language--Action policies}
    } \\
    \addlinespace[1pt]

    OpenVLA-OFT
      & 547.4 & 34.2 & 6.171 & 7.8 & 7.5
      & 503.1 & 117.0 & 6.368 & 41.0 & 35.6 \\

    PRTS
      & 459.8 & 39.5 & 3.017 & 11.7 & 11.5
      & 429.2 & 105.6 & 3.506 & 41.5 & 30.9 \\

    RDT2
      & 720.0 & 126.6 & 0.885 & \NA & \NA
      & 720.0 & 287.1 & 0.771 & \NA & \NA \\

    SpatialVLA
      & 720.0 & 135.4 & 1.608 & \NA & \NA
      & 720.0 & 390.7 & 1.074 & \NA & \NA \\

    StarVLA
      & 542.4 & 48.0 & 2.338 & 12.9 & 12.8
      & 580.5 & 144.9 & 3.871 & 29.0 & 31.4 \\

    RynnVLA-002$^{\ddagger}$
      & 673.9 & 81.9 & 1.313 & 11.1 & 11.1
      & 591.9 & 1005.2 & 0.873 & 191.5 & 145.9 \\

    EVO1
      & 542.0 & 68.5 & 2.008 & 18.4 & 15.9
      & 581.8 & 246.1 & 1.906 & 32.5 & 32.7 \\

    ABot-M0
      & 540.9 & 119.4 & 2.392 & 25.3 & 24.6
      & 580.1 & 168.8 & 2.402 & 42.2 & 41.5 \\

    InternVLA-A1.5
      & 535.5 & 157.8 & 0.914 & 67.2 & 43.6
      & 407.6 & 166.4 & 1.375 & 38.9 & 32.1 \\

    SmolVLA
      & 590.1 & 113.3 & 3.175 & 29.9 & 28.4
      & 557.0 & 242.9 & 2.360 & 55.5 & 43.4 \\

    WLA-0
      & 541.6 & 138.5 & 1.192 & 34.6 & 32.1
      & 533.4 & 235.7 & 1.522 & 27.3 & 25.6 \\

    X-VLA
      & 654.7 & 134.8 & 3.269 & 71.1 & 71.1
      & 533.6 & 223.2 & 6.201 & 38.2 & 40.2 \\

    MolmoAct2
      & 301.1 & 56.3 & 3.649 & 28.5 & 26.6
      & 213.6 & 89.0 & 2.841 & 41.7 & 40.6 \\
      
    $\pi$0.5-RLinf-SFT
      & 303.8 & 60.4 & 4.876 & 30.8 & 29.5
      & 128.9 & 30.9 & 4.870 & 23.1 & 20.0 \\

    \midrule
    \multicolumn{11}{@{}l}{
      \textit{Video--Action models}
    } \\
    \addlinespace[1pt]

    LingBot-VA
      & 720.0 & 354.7 & 2.315 & \NA & \NA
      & 625.7 & 708.1 & 1.622 & 364.9 & 227.6 \\

    \midrule
    \multicolumn{11}{@{}l}{
      \textit{World--Action models}
    } \\
    \addlinespace[1pt]

    FastWAM
      & 542.6 & 120.0 & 1.451 & 31.7 & 31.0
      & 177.7 & 51.7 & 1.459 & 10.6 & 7.1 \\

    FasterWAM
      & 542.0 & 115.5 & 2.201 & 28.3 & 27.5
      & 177.3 & 47.3 & 1.835 & 10.7 & 6.5 \\

    \bottomrule
  \end{tabular}%
  }
\end{table}
\begin{table}[H]
  \centering
  \caption{
  Execution efficiency on RoboSuite under the \textbf{Full} and \textbf{G--L}
  evaluation criteria. \textbf{Full} requires successful completion of the
  entire task, whereas \textbf{G--L} requires successful grasp and lift of the
  target object.
  For each criterion, \textbf{All trials} aggregates both successful and failed
  evaluations and reports the average number of executed steps, average
  execution time, and effective control frequency.
  \textbf{Successful only} reports the mean and median execution time computed
  only over evaluations satisfying the corresponding success criterion.
  A dash denotes that no evaluation is successful under that criterion, so
  successful-only timing statistics are unavailable; \textbf{NR} denotes a
  statistic that was not recorded.
  Lower all-trial step counts or execution times do not necessarily indicate
  higher execution efficiency, since failed evaluations may terminate early
  rather than complete the full rollout horizon.
  All-trial efficiency should therefore be interpreted jointly with the
  corresponding success counts reported in
  \Cref{tab:vla_wam_environment_results}, where method references are also
  provided.
}
  \label{tab:vla_wam_efficiency_robosuite}

  \footnotesize
  \setlength{\tabcolsep}{2.0pt}
  \renewcommand{\arraystretch}{1.05}
 
  \resizebox{\linewidth}{!}{%
  \begin{tabular}{@{}l
    S[table-format=3.1]
    S[table-format=4.1]
    S[table-format=1.3]
    S[table-format=3.1]
    S[table-format=3.1]
    S[table-format=3.1]
    S[table-format=4.1]
    S[table-format=1.3]
    S[table-format=3.1]
    S[table-format=3.1]@{}}

    \toprule
    & \multicolumn{5}{c}{Full}
    & \multicolumn{5}{c}{G--L} \\
    \cmidrule(lr){2-6}
    \cmidrule(lr){7-11}

    & \multicolumn{3}{c}{All trials}
    & \multicolumn{2}{c}{Successful only}
    & \multicolumn{3}{c}{All trials}
    & \multicolumn{2}{c}{Successful only} \\
    \cmidrule(lr){2-4}
    \cmidrule(lr){5-6}
    \cmidrule(lr){7-9}
    \cmidrule(lr){10-11}

    Method
      & {Steps}
      & {Time (s)}
      & {Hz}
      & {Mean (s)}
      & {Median (s)}
      & {Steps}
      & {Time (s)}
      & {Hz}
      & {Mean (s)}
      & {Median (s)} \\
    \midrule

    \multicolumn{11}{@{}l}{
      \textit{Vision--Language--Action policies}
    } \\
    \addlinespace[1pt]

    OpenVLA-OFT
      & 720.0 & 179.2 & 2.337 & \NA & \NA
      & 720.0 & 111.2 & 7.042 & \NA & \NA \\

    PRTS
      & 575.5 & 160.0 & 1.779 & 37.6 & 37.6
      & 561.5 & 94.5 & 3.092 & 18.2 & 18.2 \\

    RDT2
      & 720.0 & 261.7 & 0.726 & \NA & \NA
      & 720.0 & 202.5 & 0.856 & \NA & \NA \\

    SpatialVLA
      & 720.0 & 313.1 & 1.024 & \NA & \NA
      & 720.0 & 207.1 & 1.546 & \NA & \NA \\

    StarVLA
      & 200.0 & 46.7 & 3.636 & \NA & \NA
      & 720.0 & 124.9 & 3.971 & \NA & \NA \\

    RynnVLA-002$^{\ddagger}$
      & 220.0 & 192.3 & 1.478 & \NA & \NA
      & 720.0 & 590.4 & 1.462 & \NA & \NA \\

    EVO1
      & 720.0 & 128.6 & 2.516 & \NA & \NA
      & 720.0 & 128.9 & 2.343 & \NA & \NA \\

    ABot-M0
      & 576.8 & 100.1 & 3.739 & 26.2 & 26.2
      & 720.0 & 87.7 & 3.754 & \NA & \NA \\

    InternVLA-A1.5
      & 720.0 & 159.0 & 1.460 & \NA & \NA
      & 720.0 & 104.4 & 1.278 & \NA & \NA \\

    SmolVLA
      & 720.0 & 164.4 & \NR & \NA & \NA
      & 720.0 & 135.4 & 2.958 & \NA & \NA \\

    WLA-0
      & 661.2 & 139.5 & 1.717 & 97.0 & 97.0
      & 559.2 & 127.4 & 1.564 & 20.2 & 20.2 \\

    X-VLA
      & 720.0 & 101.6 & 7.382 & \NA & \NA
      & 720.0 & 103.4 & 7.431 & \NA & \NA \\

    MolmoAct2
      & 556.8 & 123.5 & 1.103 & 15.6 & 15.6
      & 404.0 & 80.9 & 2.746 & 17.3 & 17.3 \\
    
    $\pi$0.5-RLinf-SFT
      & 720.0 & 129.7 & 4.780 & \NA & \NA
      & 424.8 & 74.6 & 4.952 & 24.1 & 24.1 \\
    \midrule
    \multicolumn{11}{@{}l}{
      \textit{Video--Action models}
    } \\
    \addlinespace[1pt]

    LingBot-VA
      & 720.0 & 377.8 & 2.887 & \NA & \NA
      & 526.8 & 459.2 & 1.303 & 372.1 & 372.1 \\

    \midrule
    \multicolumn{11}{@{}l}{
      \textit{World--Action models}
    } \\
    \addlinespace[1pt]

    FastWAM
      & 720.0 & 166.5 & 1.456 & \NA & \NA
      & 720.0 & 160.5 & 1.451 & \NA & \NA \\

    FasterWAM
      & 720.0 & 142.4 & 1.520 & \NA & \NA
      & 720.0 & 149.1 & 1.820 & \NA & \NA \\

    \bottomrule
  \end{tabular}%
  }
\end{table}
\begin{table}[H]
  \centering
  \caption{
  Execution efficiency on RoboTwin 2.0 under the \textbf{Full} and
  \textbf{G--L} evaluation criteria.
  \textbf{Full} requires successful completion of the entire task, whereas
  \textbf{G--L} requires successful grasp and lift of the target object.
  For each criterion, \textbf{All trials} reports execution statistics computed
  over the complete evaluation set, including both successful and failed
  trials: \textbf{Steps}, \textbf{Time (s)}, and \textbf{Hz} characterize
  rollout length, execution time, and effective control frequency,
  respectively.
  \textbf{Successful trials} reports the mean and median execution time over
  trials satisfying the corresponding success criterion.
  A dash denotes that no trial is successful under that criterion, so
  successful-trial timing statistics are unavailable.
  Lower all-trial step counts or execution times do not necessarily indicate
  greater execution efficiency, since failed trials may terminate before task
  completion and therefore incur shorter rollouts.
  These efficiency statistics should therefore be interpreted jointly with
  the corresponding success counts reported in
  \Cref{tab:vla_wam_environment_results}, where method references are also
  provided.
}
  \label{tab:vla_wam_efficiency_robotwin}

  \footnotesize
  \setlength{\tabcolsep}{2.0pt}
  \renewcommand{\arraystretch}{1.05}

  \resizebox{\linewidth}{!}{%
  \begin{tabular}{@{}l
    S[table-format=3.1]
    S[table-format=4.1]
    S[table-format=1.3]
    S[table-format=3.1]
    S[table-format=3.1]
    S[table-format=3.1]
    S[table-format=4.1]
    S[table-format=1.3]
    S[table-format=3.1]
    S[table-format=3.1]@{}}

    \toprule
    &
    \multicolumn{5}{c}{Full}
    &
    \multicolumn{5}{c}{G--L} \\
    \cmidrule(lr){2-6}
    \cmidrule(lr){7-11}

    &
    \multicolumn{3}{c}{All trials}
    &
    \multicolumn{2}{c}{Successful trials}
    &
    \multicolumn{3}{c}{All trials}
    &
    \multicolumn{2}{c}{Successful trials} \\
    \cmidrule(lr){2-4}
    \cmidrule(lr){5-6}
    \cmidrule(lr){7-9}
    \cmidrule(lr){10-11}

    Method
      & \multicolumn{1}{c}{Steps}
      & \multicolumn{1}{c}{Time (s)}
      & \multicolumn{1}{c}{Hz}
      & \multicolumn{1}{c}{Mean (s)}
      & \multicolumn{1}{c}{Median (s)}
      & \multicolumn{1}{c}{Steps}
      & \multicolumn{1}{c}{Time (s)}
      & \multicolumn{1}{c}{Hz}
      & \multicolumn{1}{c}{Mean (s)}
      & \multicolumn{1}{c}{Median (s)} \\
    \midrule

    \multicolumn{11}{@{}l}{
      \textit{Vision--Language--Action policies}
    } \\
    \addlinespace[1pt]

    OpenVLA-OFT
      & 792.0 & 1273.4 & 3.879 & \NA & \NA
      & 675.6 & 1292.2 & 3.434 & 51.3 & 51.3 \\

    PRTS
      & 790.0 & 1620.3 & 1.370 & \NA & \NA
      & 464.5 & 654.0 & 2.139 & 78.9 & 53.0 \\

    RDT2
      & 792.0 & 1372.8 & 0.347 & \NA & \NA
      & 790.0 & 1359.1 & 0.334 & \NA & \NA \\

    SpatialVLA
      & 790.0 & 2242.1 & 0.669 & \NA & \NA
      & 739.7 & 1565.8 & 0.797 & 234.1 & 234.1 \\

    StarVLA
      & 790.0 & 708.5 & 3.938 & \NA & \NA
      & 594.7 & 621.8 & 3.321 & 96.4 & 96.4 \\

    RynnVLA-002
      & 790.0 & 1343.2 & 0.905 & \NA & \NA
      & 734.8 & 1258.9 & 0.979 & 121.3 & 121.3 \\

    EVO1
      & 806.6 & 641.9 & 1.096 & \NA & \NA
      & 535.2 & 399.3 & 1.626 & 111.7 & 61.3 \\

    ABot-M0
      & 470.0 & 245.5 & 3.610 & 144.0 & 139.6
      & 139.8 & 117.7 & 3.397 & 44.2 & 43.8 \\

    InternVLA-A1.5
      & 326.0 & 188.0 & 1.534 & 162.1 & 153.9
      & 144.8 & 78.7 & 1.511 & 40.7 & 42.5 \\

    SmolVLA
      & 790.0 & 405.9 & 2.392 & \NA & \NA
      & 574.6 & 332.6 & 2.011 & 48.1 & 38.8 \\

    WLA-0
      & 368.0 & 265.6 & 1.295 & 206.3 & 202.4
      & 81.4 & 65.2 & 1.318 & 65.2 & 54.0 \\

    X-VLA
      & 798.0 & 481.6 & 7.936 & \NA & \NA
      & 790.0 & 493.5 & 7.557 & \NA & \NA \\

    MolmoAct2
      & 792.0 & 670.4 & 3.983 & \NA & \NA
      & 790.0 & 359.7 & 0.871 & \NA & \NA \\

    $\pi$0.5-RLinf-SFT
      & 790.0 & 616.1 & 5.029 & \NA & \NA
      & 666.8 & 483.8 & 5.042 & 32.6 & 32.6 \\

    \midrule

    \multicolumn{11}{@{}l}{
      \textit{Video--Action models}
    } \\
    \addlinespace[1pt]

    LingBot-VA
      & 344.5 & 203.8 & 0.948 & 150.0 & 159.3
      & 138.4 & 168.4 & 0.331 & 101.1 & 103.8 \\

    \midrule

    \multicolumn{11}{@{}l}{
      \textit{World--Action models}
    } \\
    \addlinespace[1pt]

    FastWAM
      & 284.8 & 164.1 & 0.835 & 164.1 & 153.6
      & 73.2 & 40.0 & 1.453 & 40.0 & 38.7 \\

    FasterWAM
      & 431.2 & 239.8 & 1.173 & 178.2 & 179.3
      & 80.1 & 44.3 & 1.863 & 44.3 & 45.6 \\

    \bottomrule
  \end{tabular}%
  }
\end{table}
}

\endgroup

\section{Experiment Details}\label{app:experimental-setup}

\subsection{Benchmark Tasks}\label{app:benchmark-details}
The evaluation suite comprises $100$ tabletop-manipulation tasks across three benchmarks with different scene and control requirements: LIBERO-Pro~\citep{LIBERO-Pro} (single-arm tasks under instruction-redirection~\textbf{T} and position-swap~\textbf{S} perturbations), RoboSuite~\citep{RoboSuite} (single-arm pick--place in clean scenes), and RoboTwin~2.0~\citep{RoboTwin} (long-horizon bimanual manipulation).
The task pool contains $80$ LIBERO-Pro, $4$ RoboSuite, and $16$ RoboTwin~2.0 tasks. Coordinator training uses scene seed~$0$ for these tasks, while evaluation uses seeds~$1$--$10$. This split tests generalization to new scene seeds within the same task pool.
LIBERO-Pro columns in~\mainCref{tab:main_exp} follow the eight T/S splits (Spat, Obj, Goal, and L10), each aggregating $10$ tasks $\times$ $10$ seeds; RoboSuite and RoboTwin columns aggregate their full task pools under the same seed range.
Mobile-manipulation suites such as BEHAVIOR~\citep{BEHAVIOR} and RoboCasa~\citep{RoboCasa} fall outside the \pschema{} tabletop-manipulation scope and are not included.

\paragraph{Cross-agent and cross-embodiment sensitivity.}
The study described in Takeaway~5 uses a fixed $3\times4$ backend\textendash{}embodiment grid on four RoboTwin~2.0 task instances:
\texttt{blocks\_ranking\_rgb} (arrange the red, green, and blue blocks),
\texttt{place\_cans\_plasticbox} (place the cans into the plastic box),
\texttt{place\_bread\_basket} (place the bread into the basket), and
\texttt{stack\_blocks\_two} (stack two blocks).
Claude~Code and Aloha\textendash{}AgileX are the default backend and embodiment, respectively; the comparison swaps in Cursor or Codex, and Franka\textendash{}Panda, ARX\textendash{}X5, or UR5\textendash{}WSG, while retaining the same task instructions, \pschema{} stage contracts, and typed API interfaces.
Cursor and Claude~Code both use Claude~Opus~4.8 as the underlying generator, while Codex uses GPT-5.5.
The backend sweep keeps the API library and trained coordinator fixed, whereas each embodiment uses its corresponding native adapter and compatible low-level controller under the shared API contract; client and decoding details are given in~\Cref{app:coding-agent-details}.
The supplementary videos provide one demonstration for each task\textendash{}embodiment pair, giving 16 clips organized as \texttt{<task>/<embodiment>.mp4}. These clips illustrate execution on the different embodiments, while~\mainCref{fig:agent-backend} reports the aggregate success rates.

\subsection{Baseline Implementations}\label{app:baseline-details}

All methods in~\mainCref{tab:main_exp} share the single-episode evaluation protocol of~\Cref{app:evaluation-metrics}: seeds~$1$--$10$, one trial per seed, terminal SR only, and \emph{no} global cross-episode memory.
For implementations that support memory across episodes, retrieval over prior tasks, or persistent tip stores, we disable those channels to keep the available experience limited to the current episode.
RoboTwin columns marked ``/'' indicate embodiment mismatch for single-arm VLA checkpoints that are incompatible with the bimanual RoboTwin controllers.

\paragraph{End-to-end policies.}
$\pi_0$~\citep{pi0} and $\pi_{0.5}$~\citep{pi05} run as frozen open-loop policies under their released checkpoints for the LIBERO / RoboSuite control stacks; they are not re-fine-tuned on our coordinator training pool.
FastWAM~\citep{FastWAM} and FasterWAM~\citep{FasterWAM} likewise use released world-action checkpoints as full-episode controllers (not as \performendtoend{} routes).
In~\algo{}, the retained $\pi_{0.5}$ and FastWAM routes appear only inside typed \performendtoend{} calls at local skill granularity (\Cref{app:perform-vla}).

\paragraph{Code-as-policy.}
CaP~\citep{CaP} and CaP-X~\citep{Cap-X} follow their public code-generation loops over language-conditioned robot APIs; we evaluate them under the shared episode horizon and without their optional cross-episode repair memory.
RATS~\citep{RATS} and ASPIRE~\citep{ASPIRE} keep their published planner / skill-library interfaces, again with global memory disabled.
These baselines use their published execution-selection rules and do not receive the learned $Q_\theta$ family coordinator.

\paragraph{VLA with harness.}
Pigey~\citep{Pigey} and VLS~\citep{VLS} use their public harness wrappers around a frozen generative policy, with early-stop and reward-steering settings taken from the authors' default configs under the shared single-episode protocol.
Harness-VLA~\citep{HarnessVLA} is evaluated by us using the RLinf-released \texttt{pi05\_libero130\_fullshot} $\pi_{0.5}$-RLinf-SFT checkpoint~\citep{RLinf}, exposed through the \textsc{vla\_act} primitive; harness component ablations appear in~\Cref{app:harness-vla-ablation}.
Controlled routing ablations of~\algo{} (\algo{}-UO, \algo{}-TF, \algo{}-UA) keep the same coding agent and API library and change only coordinator training or family allocation, as listed in~\mainCref{tab:paired-supervision,tab:ablation-analysis}.

\subsection{Success and Runtime Metrics}\label{app:evaluation-metrics}
This subsection defines the success-rate and runtime measures used in the main evaluation. Each table specifies its task pool, and comparisons within a table use the same success criterion and timing convention unless stated otherwise.
The Harness-VLA ablation in~\Cref{app:harness-vla-ablation} uses a separate task and seed schedule and reports success among valid completions, together with execution coverage.

\paragraph{Success-rate denominator.}
An evaluation episode is one \texttt{(task, scene seed)} pair under seeds~$1$--$10$.
The binary success indicator is the terminal task reward $r\in\{0,1\}$ emitted by the suite's registered evaluator when the episode ends by task completion, unrecoverable failure, empty $\mathcal{C}_i$, or the horizon $T$; partial stage progress never counts as success.
The episode-level success rate (SR) is therefore
\[
\mathrm{SR}_{\mathrm{ep}}=\frac{1}{N}\sum_{n=1}^{N}\mathbf{1}[r_n=1],
\]
where $N$ is the number of evaluated episodes in the stated pool (for example, $N{=}100$ for a $10$~tasks $\times$ $10$~seeds column in~\mainCref{tab:main_exp}, and $N{=}1{,}000$ for the full $100$~tasks $\times$ $10$~seeds suite).
In the main evaluation, episodes that abort remain in the denominator, including crashes, timeouts, and failures caused by an empty admissible-family set.

\paragraph{Task- and suite-level weighting.}
To distinguish equal weighting of tasks from pooling individual episodes, we define two averages.
\emph{Task macro-average} first computes a per-task SR over that task's seeds, then averages those task rates with equal weight:
\[
\mathrm{SR}_{\mathrm{macro}}=\frac{1}{|\mathcal{T}|}\sum_{t\in\mathcal{T}}\mathrm{SR}_{\mathrm{ep}}(t).
\]
\emph{Episode-weighted} (micro) average is $\mathrm{SR}_{\mathrm{ep}}$ over the pooled episodes and therefore weights tasks by their episode counts.
Because every task in the main evaluation has the same number of seeds, the task macro-average and episode-weighted average coincide, both within each suite and across all $1{,}000$ episodes. They differ only when tasks contribute unequal numbers of episodes.
An unweighted average of the three suite-level rates is a different measure. It gives equal weight to LIBERO-Pro, RoboSuite, and RoboTwin despite their unequal task counts, whereas pooling all episodes gives each task equal weight under this protocol.
We therefore identify whether each cross-suite result is an equal-weight suite average or a pooled episode average. Main-text columns use the episode-weighted measure specified in their captions.

\paragraph{Execution-time scope.}
Reported times are wall-clock seconds on the evaluation host unless a table names a different clock.
\emph{End-to-end episode time} starts at the post-reset observation used to build the task plan and stops when the episode terminates under the success predicate above; simulator reset and offline model loading are excluded.
Averages over all trials include failures; successful-trial means or medians are computed only on episodes with $r{=}1$ and are marked undefined when that set is empty, matching~\Cref{tab:vla_wam_efficiency_robotwin}.

\subsection{Hyperparameters and Experimental Settings}\label{app:hyperparameters}
\Cref{tab:hyperparameters} lists the reward weights, search and learning settings, and step budgets used by \algo{}.
We distinguish four budgets because they count different parts of execution:
\emph{environment / control steps} advance the physics or low-level controller inside an API;
an \emph{API step budget} caps how long an atomic call may retain control before returning success or a typed failure;
\emph{semantic decision steps} count \pschema{} boundaries at which the coordinator may select a family;
and the MCTS budget $K$ counts \emph{completed suffix simulations} per search root, not environment steps or semantic decisions.
Root-collection and rollout suffixes use a fixed family-sampling rule with support on every admissible family; tree Selection uses UCB (\cref{eq:ucb}) instead.
The main experiments use $K{=}20$. The RoboSuite ablation compares $K\in\{5,10,20,50\}$ and shows diminishing success-rate gains beyond $K{=}20$ (\mainCref{fig:explore-budget}).
\algo{}-UA uses uniform family allocation under the same $K$ values.

\begin{table}[htbp]
  \centering
  \footnotesize
  \setlength{\tabcolsep}{8pt}
  \caption{Hyperparameters of \algo{}.}
  \label{tab:hyperparameters}
  \renewcommand{\arraystretch}{1.12}
  \begin{tabular}{@{}lcl@{}}
    \toprule
    Parameter & Symbol & Setting \\
    \midrule
    Discount factor & $\gamma$ & $1$ \\
    Perception reward weight & $w_{\mathrm p}$ & $0.2$ \\
    Manipulation reward weight & $w_{\mathrm m}$ & $0.3$ \\
    Task-success reward weight & $w_{\mathrm t}$ & $0.5$ \\
    Smooth-$L_1$ threshold & \textemdash{} & $1$  \\
    \midrule
    Neighboring-root draws & $S$ & $4$ \\
    Neighboring-time window & $\Delta$ & $10$ saved control steps \\
    UCB coefficient & $c_{\mathrm{ucb}}$ & $1.0$ \\
    Suffix-simulation budget & $K$ & $20$ \\
    Episode horizon & $T$ & $100$ decision transitions \\
    \midrule
    Q-learning updates & $M$ & $10{,}000$ \\
    Learning rate & $\eta$ & $10^{-5}$ \\
    Target-copy interval & $J$ & $500$ updates \\
    \midrule
    Translational pose threshold & $\epsilon_p$ & $0.02\,\mathrm{m}$ \\
    Rotational pose threshold & $\epsilon_R$ & $\pi/12\,\mathrm{rad}$ \\
    Atomic API step budget & \textemdash{} & $100$ control steps \\
    Block wall-clock timeout & \textemdash{} & $30\,\mathrm{s}$ \\
    Code-repair attempts & $R_{\mathrm{repair}}$ & $2$ \\
    Coding-agent history window & $H_{\mathrm{code}}$ & $4$ full stage transitions \\
    \bottomrule
  \end{tabular}
\end{table}

\subsection{Coding-Agent Backends}\label{app:coding-agent-details}
The coding agent $\codepolicy$ is frozen before coordinator training and is shared by~\algo{} and the controlled routing baselines.
Cursor, Claude~Code, and Codex are the client toolchains through which the coding models are accessed.
In the reported runs, Cursor and Claude~Code both use Claude~Opus~4.8, while Codex uses GPT-5.5. We fix the client/CLI versions and decoding settings and record them in the run logs.
Sensitivity results in~\mainCref{fig:agent-backend} swap only this backend while keeping the API library,~\pschema{} contracts, and $Q_\theta$ fixed.

\subsection{Code-Execution Protocol}\label{app:code-execution-protocol}

\paragraph{Execution sandbox, budgets, and retries.}
Generated blocks run in a restricted Python process that can import only the registered API library and allowlisted Python standard-library helpers.
Each block inherits an API-call budget, a control-step budget, and a $30\,\mathrm{s}$ wall-clock timeout (see~\Cref{tab:hyperparameters}).
Syntax errors, undefined names, disallowed imports, and API schema violations trigger an automatic repair round that appends the error text and regenerates under the same family constraint, up to $R_{\mathrm{repair}}{=}2$ attempts; exhausted repairs mark the branch as failed for~\scb{} or terminate the deployment stage according to~\Cref{tab:stage-contracts}.
Runtime exceptions raised inside an allowed API are returned as typed failure codes to the agent rather than as raw simulator dumps.

\paragraph{Episode reset versus memory.}
Between evaluation episodes (and between independent training trajectories), the coding-client session is reset: conversation state, tool caches, and any product-level memory are cleared so that episode~$n{+}1$ cannot read traces from episode~$n$.
Within an episode, feedback is limited to the observable channel already present in $x_i$: typed API returns, postcondition bits, failure codes, timeouts, and the truncated history $h_i$.
We do \emph{not} enable cross-episode product memory, retrieval over prior tasks, or persistent ``global tips'' stores for main results; any ablation that reintroduces such memory is labeled separately and is not part of the default~\algo{} protocol.
Family selection at deployment is supplied by $Q_\theta$; the coding agent does not choose $c_i$, and prompts never reveal the unchosen family's counterfactual outcome.

\subsection{Computation and Software}\label{app:reproduction-details}

The reported coordinator run used four NVIDIA A100 GPUs with 80~GB memory each.
Fitting $Q_\theta$ took approximately six hours, corresponding to approximately 24 GPU-hours ($4\times6\,\mathrm{h}$). Trajectory and counterfactual data collection took approximately one additional day on the same hardware pool.
Data collection therefore accounts for most of the wall-clock time and must be included when reporting the training cost.
Supplementary qualitative clips for the cross-embodiment study follow the naming \texttt{<task>/<embodiment>.mp4} described in~\Cref{app:benchmark-details}.

\section{Additional Empirical Results}\label{app:supplementary-results}

\subsection{Ablation of Harness-VLA}\label{app:harness-vla-ablation}
\paragraph{Experimental setup.}
We evaluate the harness components in two stages. All six configurations in
\Cref{tab:harness_ablation_components} use the same $\pi_{0.5}$ SFT backbone and
a common neutral interface prompt. We first screen them on 24 tasks with
seed~42, then evaluate Full, w/o Memory, and w/o Memory \& SAM3 on 80 tasks
with seeds $\{41,42,43\}$.
Removing Memory also disables the seed-0 reference, so we use \emph{memory
package} to denote these two components together. The w/o Reference variant
removes only the reference while retaining Memory, allowing us to examine
that component separately in the screening stage.

\begin{table*}[t]
  \centering
  \caption{
    Component configurations for the Harness VLA ablation study.
    All variants use the same $\pi_{0.5}$ SFT backbone.
  }
  \label{tab:harness_ablation_components}
  \small
  \setlength{\tabcolsep}{6pt}
  \renewcommand{\arraystretch}{1.12}
  \resizebox{\textwidth}{!}{%
    \begin{tabular}{@{}lcccccc@{}}
      \toprule
      Method
      & Planner
      & Memory
      & \shortstack{Seed-0\\reference}
      & SAM3
      & \shortstack{Scripted\\hybrid}
      & VLA \\
      \midrule
      A: Full Harness VLA
      & $\checkmark$ & $\checkmark$ & $\checkmark$
      & $\checkmark$ & $\checkmark$ & $\pi_{0.5}$ SFT \\

      B: w/o Memory
      & $\checkmark$ & $\times$ & $\times$
      & $\checkmark$ & $\checkmark$ & $\pi_{0.5}$ SFT \\

      C: w/o Reference
      & $\checkmark$ & $\checkmark$ & $\times$
      & $\checkmark$ & $\checkmark$ & $\pi_{0.5}$ SFT \\

      D: w/o SAM3
      & $\checkmark$ & $\checkmark$ & $\checkmark$
      & $\times$ & $\checkmark$ & $\pi_{0.5}$ SFT \\

      E: w/o Memory \& SAM3
      & $\checkmark$ & $\times$ & $\times$
      & $\times$ & $\checkmark$ & $\pi_{0.5}$ SFT \\

      F: Bare SFT
      & $\times$ & $\times$ & $\times$
      & $\times$ & $\times$ & $\pi_{0.5}$ SFT \\
      \bottomrule
    \end{tabular}%
  }
\end{table*}

\paragraph{Single-seed screening.}
Full succeeds on 15 of 24 tasks (62.5\%;
\Cref{tab:harness_ablation_screening}). Removing both the memory package and
SAM3 produces the largest observed decrease, to 37.5\%. This is a drop of
25.0 percentage points (pp), with a 95\% task-bootstrap confidence interval
of $[4.2,45.8]$ pp.
Removing the memory package alone, the seed-0 reference alone, or SAM3 alone
reduces success by 16.7, 12.5, and 4.2 pp, respectively, although all three
confidence intervals include zero. The larger drop from joint removal does
not by itself establish a positive interaction. The estimated additive
interaction, $\mathrm{SR}_{A}-\mathrm{SR}_{B}-\mathrm{SR}_{D}+\mathrm{SR}_{E}$,
is $-4.2$ pp and provides no evidence of a positive additive interaction.

Bare SFT achieves 41.7\%, which is 20.8 pp below Full. Because this variant
also removes the planner and scripted hybrid execution, its comparison with
Full measures the contribution of the overall harness rather than the planner
alone.

\begin{table*}[t]
  \centering
  \caption{
    Paired ablation screening on 24 tasks using seed 42.
    Drop is defined as the success rate of Full minus that of
    the corresponding variant, in percentage points (pp).
    Confidence intervals are 95\% task-bootstrap intervals
    for the paired drop.
    S$\rightarrow$F counts tasks solved by Full but failed by the ablated variant; F$\rightarrow$S counts the reverse.
    All 24 episodes completed without errors in each variant.
  }
  \label{tab:harness_ablation_screening}
  \small
  \setlength{\tabcolsep}{6pt}
  \renewcommand{\arraystretch}{1.12}
  \resizebox{\textwidth}{!}{%
    \begin{tabular}{@{}lrrrcrr@{}}
      \toprule
      Method
      & \shortstack{Success /\\valid}
      & \shortstack{Success\\rate (\%)}
      & \shortstack{Drop\\(pp)}
      & \shortstack{95\% CI\\(pp)}
      & S$\rightarrow$F
      & F$\rightarrow$S \\
      \midrule
      A: Full Harness VLA
      & 15/24 & \textbf{62.5} & -- & -- & -- & -- \\

      B: w/o Memory
      & 11/24 & 45.8 & 16.7 & $[-8.3,\ 41.7]$ & 7 & 3 \\

      C: w/o Reference
      & 12/24 & 50.0 & 12.5 & $[-12.5,\ 37.5]$ & 6 & 3 \\

      D: w/o SAM3
      & 14/24 & 58.3 & 4.2 & $[-8.3,\ 20.8]$ & 2 & 1 \\

      E: w/o Memory \& SAM3
      & 9/24 & 37.5 & \textbf{25.0} & $[4.2,\ 45.8]$ & 7 & 1 \\

      F: Bare SFT
      & 10/24 & 41.7 & 20.8 & $[0.0,\ 41.7]$ & 6 & 1 \\
      \bottomrule
    \end{tabular}%
  }
\end{table*}

\paragraph{Multi-seed confirmation.}
\Cref{tab:harness_ablation_confirmation} separates execution errors from
completed episodes in which the task fails. Among valid completions, Full
achieves 56.6\% success, compared with 16.9\% for w/o Memory and 20.9\% for
w/o Memory \& SAM3.
Because execution errors leave different sets of valid episodes for each
variant, we also compare variants on task--seed pairs that completed under
both configurations. On these paired subsets, removing the memory package
yields a signed difference of $-38.9$ pp ($n=113$, 95\% CI
$[-50.0,-27.4]$ pp), while removing both the memory package and SAM3 yields
$-35.8$ pp ($n=109$, 95\% CI $[-48.2,-23.4]$ pp).
Both intervals exclude zero, supporting a contribution from the memory
package on the jointly completed episodes.

\paragraph{Scope and limitations.}
Service-connection errors leave only 47.1--49.2\% of the 240 scheduled
episodes per configuration with valid completions. The paired estimates
therefore describe those jointly completed episodes and cannot establish the
effect over the full scheduled set. We report execution coverage separately
and do not count these service errors as model failures in this ablation.
The confirmation stage also omits w/o SAM3, so it does not independently
measure the standalone contribution of SAM3.
The historical whole-system result of 65/80 (81.25\%) used a different prompt
and protocol and is therefore not directly comparable with Full in this study.
\begin{table*}[b]
  \centering
  \caption{
    Multi-seed confirmation on 80 tasks with seeds
    $\{41,42,43\}$, yielding 240 scheduled episodes per variant.
    Valid SR is the number of successful episodes divided by the number of successful and failed episodes, so execution errors are excluded.
    Coverage is the fraction of the 240 scheduled episodes that completed without errors.
    Paired differences compare each variant against Full on task--seed pairs that completed without errors in both conditions:
    $\Delta=\mathrm{SR}_{\mathrm{variant}}-\mathrm{SR}_{\mathrm{Full}}$ on that jointly valid subset of size $n_{\mathrm{pair}}$, which need not equal the gap between the separately reported valid success rates.
  }
  \label{tab:harness_ablation_confirmation}
  \small
  \setlength{\tabcolsep}{4pt}
  \renewcommand{\arraystretch}{1.12}
  \resizebox{\textwidth}{!}{%
    \begin{tabular}{@{}lrrrrrrrrc@{}}
      \toprule
      Method
      & Success
      & Failure
      & Error
      & \shortstack{Valid /\\scheduled}
      & \shortstack{Valid SR\\(\%)}
      & \shortstack{Coverage\\(\%)}
      & $n_{\mathrm{pair}}$
      & \shortstack{Paired $\Delta$\\(pp)}
      & \shortstack{95\% CI\\(pp)} \\
      \midrule
      A: Full Harness VLA
      & 64 & 49 & 127
      & 113/240 & \textbf{56.6} & 47.1
      & -- & -- & -- \\

      B: w/o Memory
      & 20 & 98 & 122
      & 118/240 & 16.9 & 49.2
      & 113 & $\mathbf{-38.9}$ & $[-50.0,\ -27.4]$ \\

      E: w/o Memory \& SAM3
      & 24 & 91 & 125
      & 115/240 & 20.9 & 47.9
      & 109 & $\mathbf{-35.8}$ & $[-48.2,\ -23.4]$ \\
      \bottomrule
    \end{tabular}%
  }
\end{table*}

\subsection{Visualizations}\label{app:visualizations}
\begin{figure}[htbp]
  \centering
  \includegraphics[width=0.75\linewidth]{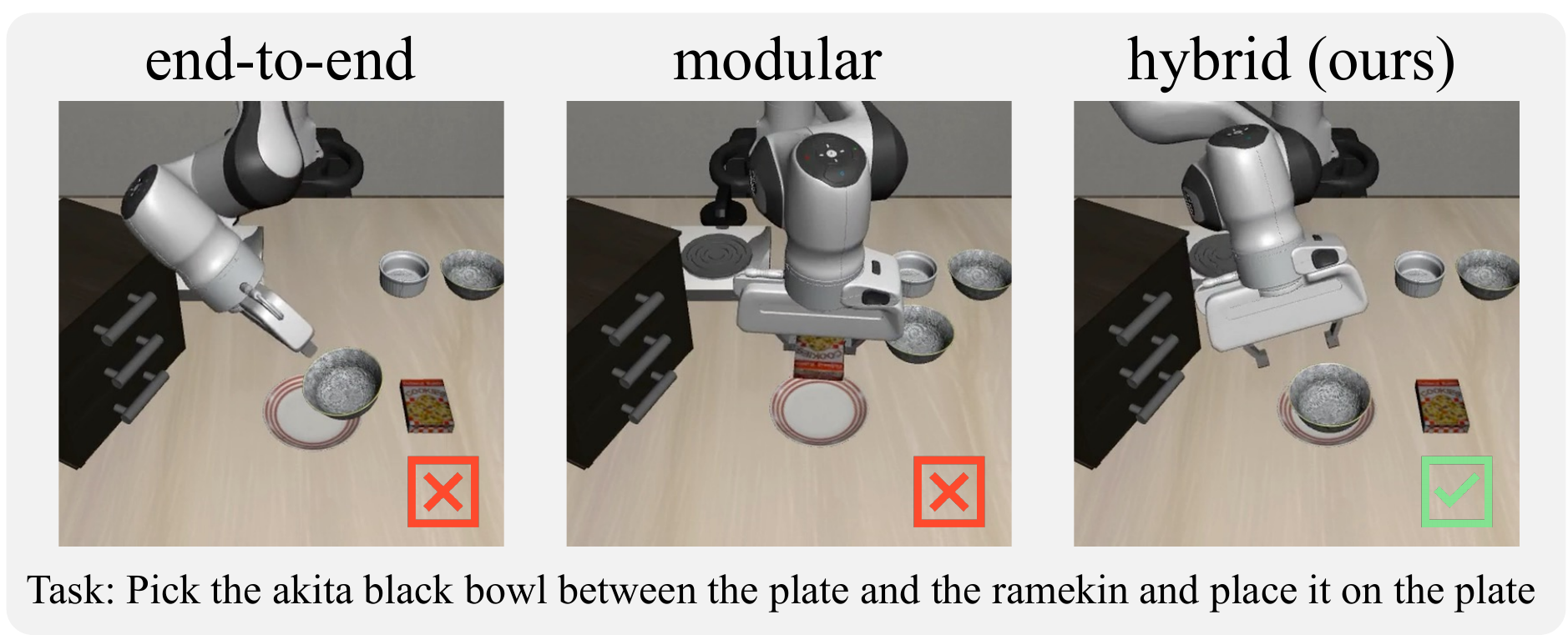}
  \caption{
    Additional LIBERO-Pro comparison on
    ``Pick the akita black bowl between the plate and the ramekin and place it on the plate,''
    complementing~\mainCref{fig:ablation-hybrid}.
    \algo{}-EO approaches with a misaligned gripper and fails to secure the bowl.
    \algo{}-MO completes a grasp but selects the wrong object (the cookie box).
    \algo{} grounds the spatially described bowl and finishes the pick-and-place,
    succeeding by switching policy families across \pschema{} boundaries.
  }
  \label{fig:ablation-hybrid-app}
\end{figure}

\Cref{fig:ablation-hybrid-app} shows a further LIBERO-Pro rollout under spatial-language perturbation, extending the cases in~\mainCref{fig:ablation-hybrid}.
As before, \algo{}-EO and \algo{}-MO fail in complementary ways, whereas \algo{} completes the task by switching families at \pschema{} boundaries.

\subsection{Failure Modes}\label{app:qualitative-analysis}
\begin{figure}[htbp]
  \centering
  \includegraphics[width=.75\linewidth]{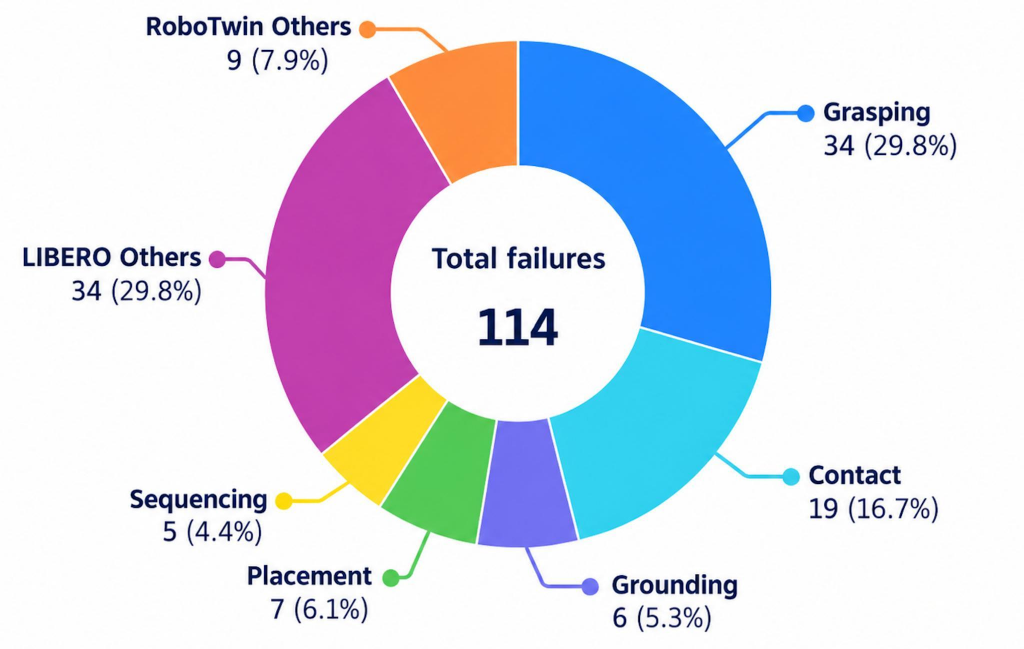}
  \caption{
    Failure modes of the unsuccessful rollouts.
    Grasping instability is the largest identified failure mode (29.8\%),
    followed by constrained contact or insertion failures (16.7\%).
    Grounding, placement, and multi-object sequencing account for 5.3\%,
    6.1\%, and 4.4\%, respectively.
    \textit{Others} (37.7\%) denotes failed rollouts for which the saved evaluation
    records are insufficient to identify a specific cause
    (29.8\% from LIBERO-Pro and 7.9\% from RoboTwin).
  }
  \label{fig:failure_modes}
\end{figure}

\Cref{fig:failure_modes} summarizes the failures observed in LIBERO-Pro and
RoboTwin. Grasping and object-holding instability is the largest identified
category, accounting for 29.8\% of failed rollouts. Constrained contact,
insertion, and mechanism actuation account for another 16.7\%.
Together, these categories represent 46.5\% of all failed rollouts, indicating
that low-level interaction and contact-rich execution are major sources of
failure. Target grounding (5.3\%), transport and placement (6.1\%), and
multi-object sequencing (4.4\%) account for smaller identified shares.

However, 37.7\% of failures remain in \textit{Others} because the saved
evaluation records lack the detailed traces needed to distinguish perception,
planning, control, and success-check errors. This limits how strongly the
identified categories can characterize the full failure distribution.
The categories summarize the observed outcomes. During deployment, recovery
and termination instead follow the typed API failures, timeouts, and stage
conditions defined in~\Cref{tab:stage-contracts,alg:roboaware-deploy}.

\end{document}